%% file: main.tex
\documentclass[acmtog]{acmart}

\usepackage{xspace}
\newcommand{\nickname}{AvatarCLIP}
\newcommand{\eg}{\textit{e.g.}\xspace}
\newcommand{\ie}{\textit{i.e.}\xspace}

\definecolor{shape}{RGB}{10,147,157}
\definecolor{appearance}{RGB}{238, 155, 0}
\definecolor{motion}{RGB}{174, 32, 17}

\usepackage{color}
\usepackage{xcolor}

\setcopyright{acmcopyright}\acmJournal{TOG}
\acmYear{2022}\acmVolume{41}\acmNumber{4}\acmArticle{161}\acmMonth{7} \acmDOI{10.1145/3528223.3530094}
\acmSubmissionID{297}

\begin{document}

\title{AvatarCLIP: Zero-Shot Text-Driven Generation and Animation of 3D Avatars}

\author{Fangzhou Hong}
\authornote{Both authors contributed equally to this research.}
\email{fangzhou001@ntu.edu.sg}
\orcid{0000-0003-2412-1141}
\affiliation{%
  \institution{S-Lab, Nanyang Technological University}
  \country{Singapore}
}
\author{Mingyuan Zhang}
\authornotemark[1]
\email{mingyuan001@ntu.edu.sg}
\orcid{0000-0001-8212-715X}
\affiliation{%
  \institution{S-Lab, Nanyang Technological University}
  \country{Singapore}
}
\author{Liang Pan}
\email{liang.pan@ntu.edu.sg}
\orcid{0000-0003-1821-4296}
\affiliation{%
  \institution{S-Lab, Nanyang Technological University}
  \country{Singapore}
}
\author{Zhongang Cai}
\email{caizhongang@sensetime.com}
\orcid{0000-0002-1810-3855}
\affiliation{%
  \institution{S-Lab, Nanyang Technological University}
  \country{Singapore}
}
\affiliation{%
  \institution{SenseTime Research}
  \country{China}
}
\author{Lei Yang}
\email{yanglei@sensetime.com}
\orcid{0000-0002-0571-5924}
\affiliation{%
  \institution{SenseTime Research}
  \country{China}
}
\author{Ziwei Liu}
\authornote{corresponding author}
\email{ziwei.liu@ntu.edu.sg}
\orcid{0000-0002-4220-5958}
\affiliation{%
  \institution{S-Lab, Nanyang Technological University}
  \country{Singapore}
}
\renewcommand{\shortauthors}{Fangzhou and Mingyuan, et al.}

\input{sections/00_abstract}

\ccsdesc[500]{Computing methodologies~Rendering}
\ccsdesc[300]{Computing methodologies~Animation}
\ccsdesc[100]{Computing methodologies~Shape modeling}

\keywords{Zero-Shot Generation, Text-Driven Generation, 3D Avatar Generation, 3D Avatar Animation}

\input{figures/teaser}

\maketitle

\input{sections/01_introduction}
\input{sections/02_related_work}
\input{sections/03_methods}

\input{sections/04_experiments}
\input{sections/05_conclusion}

\begin{acks}
This study is supported by NTU NAP, MOE AcRF Tier 2 (T2EP20221-0033), and under the RIE2020 Industry Alignment Fund – Industry Collaboration Projects (IAF-ICP) Funding Initiative, as well as cash and in-kind contribution from the industry partner(s).
\end{acks}

\clearpage

\bibliographystyle{ACM-Reference-Format}
\bibliography{acmart}

\end{document}

%% file: sections/00_abstract.tex
\begin{abstract}
    
    3D avatar creation plays a crucial role in the digital age. However, the whole production process is prohibitively time-consuming and labor-intensive. To democratize this technology to a larger audience, we propose \nickname{}, a zero-shot text-driven framework for 3D avatar generation and animation. Unlike professional software that requires expert knowledge, \nickname{} empowers layman users to customize a 3D avatar with the desired shape and texture, and drive the avatar with the described motions using solely natural languages. Our key insight is to take advantage of the powerful vision-language model CLIP for supervising neural human generation, in terms of 3D geometry, texture and animation. Specifically, driven by natural language descriptions, we initialize 3D human geometry generation with a shape VAE network. Based on the generated 3D human shapes, a volume rendering model is utilized to further facilitate geometry sculpting and texture generation. Moreover, by leveraging the priors learned in the motion VAE, a CLIP-guided reference-based motion synthesis method is proposed for the animation of the generated 3D avatar. Extensive qualitative and quantitative experiments validate the effectiveness and generalizability of \nickname{} on a wide range of avatars. Remarkably, \nickname{} can generate unseen 3D avatars with novel animations, achieving superior zero-shot capability. Codes are available at \textcolor{blue}{\url{https://github.com/hongfz16/AvatarCLIP}}.
    
\end{abstract}

%% file: figures/teaser.tex
\begin{teaserfigure}
  \includegraphics[width=\textwidth]{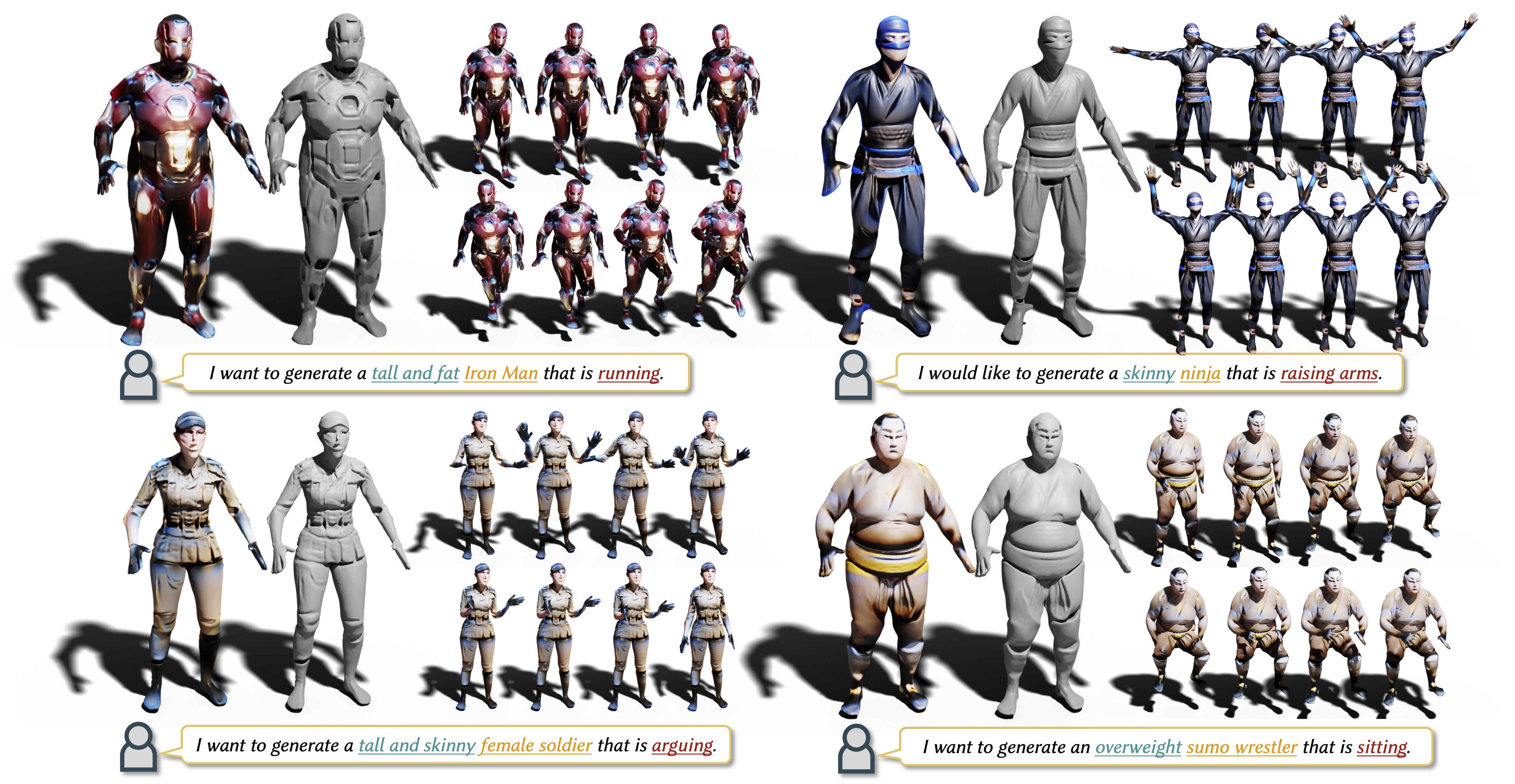}
  \caption{In this work, we present \nickname{}, a novel zero-shot text-driven 3D avatar generation and animation pipeline. Driven by natural language descriptions of the desired \textcolor{shape}{shape}, \textcolor{appearance}{appearance} and \textcolor{motion}{motion} of the avatar, \nickname{} is capable of robustly generating 3D avatar models with vivid texture, high-quality geometry and reasonable motions.}
  \Description{Teaser of this paper.}
  \label{fig:teaser}
\end{teaserfigure}

%% file: sections/01_introduction.tex
\section{Introduction}
\label{sec:intro}

Creating digital avatars has become an important part in movie, game, and fashion industries. The whole process includes creating shapes of the character, drawing textures, rigging skeletons, and driving the avatar with the captured motions. Each step of the process requires many specialists that are familiar with professional software, large numbers of working hours, and expensive equipment, which are only affordable by large companies. The good news is that recent progress in academia, such as large-scale pre-trained models~\cite{radford2021learning} and advanced human representations~\cite{loper2015smpl} are making it possible for this complicated work to be available to small studios and even reach out to the mass crowd. In this work, we take a step further and propose \nickname{}, which is capable of generating and animating 3D avatars solely from natural language descriptions as shown in Fig.~\ref{fig:teaser}.

Previously, there are several similar efforts, \eg avatar generation, and motion synthesis, towards this vision. For avatar generation, several attempts have been made in 2D-GAN-based human generation~\cite{sarkar2021humangan, lewis2021tryongan} and 3D neural human generation~\cite{grigorev2021stylepeople, zhang20213d}. However, most of them either compromise the generation quality and diversity or have limited control over the generation process. Not to mention, most generated avatars cannot be easily animated. As for motion synthesis, impressive progresses have been made in recent years. They are able to generate motions conditioned on action classes~\cite{petrovich2021action, guo2020action2motion} or motion trajectories~\cite{kania2021trajevae,ling2020character}. However, they typically require paired data for fully supervised training, which not only limits the richness of the generated motions but also makes the generation process less flexible. There still exists a considerable gap between existing works and the vision of making digital avatar creation manageable for the mass crowd.

To skirt the complicated operations, natural languages could be used as a user-friendly control signal for convenient 3D avatars generation and animation.
However, there are no existing high-quality avatar-text datasets to support supervised text-driven 3D avatar generation.
As for animating 3D avatars, a few attempts~\cite{ghosh2021text, tevet2022motionclip} have been made towards text-driven motion generation by leveraging a motion-text dataset. Nonetheless, restricted by the scarcity of paired motion-text data, those fully supervised methods have limited generalizability.

Fortunately, recent advances in vision-language \textcolor{black}{models} pave the way toward zero-shot text-driven generation. CLIP~\cite{radford2021learning} is a vision-language pre-trained model trained with large-scale image-text pairs. Through direct supervision on images, CLIP shows great success in zero-shot text-guided image generation~\cite{ramesh2021zero}. Inspired by this thread of works, we choose to take advantage of the powerful CLIP to achieve zero-shot text-driven generation and animation of 3D avatars. However, neither the 3D avatar nor motion sequences can be directly supervised by CLIP. Major challenges exist in both static avatar generation and motion synthesis.

For static 3D avatar generation, the challenges lie in three aspects, namely texture generation, geometry generation, and the ability to be animated.
Inspired by the recent advances in neural rendering~\cite{tewari2021advances}, the CLIP supervision can be applied to the rendered images to guide the generation of an implicit 3D avatar, which facilitates the generation of avatar textures.
Moreover, to speed up the optimization, we propose to initialize the implicit function based on a template human mesh. At optimization time, we also use the template mesh constraint to control the overall shape of the implicit 3D avatar.
To adapt to the modern graphics pipeline and, more importantly, to be able to be animated later, we need to extract meshes from generated implicit representations. Therefore, other than the texture, it is also desirable to generate high-quality geometry. To tackle the problem, the key insight is that when inspecting the geometry of 3D models on the computer, users usually turn off texture shading to get texture-less \textcolor{black}{renderings}, which can directly reveal the geometry. Therefore, we propose to randomly cast light on the surface of the implicit 3D avatar to get texture-less renderings, upon which the CLIP supervision is applied.
Last but not least, to make the generated implicit 3D avatar animatable, we propose to leverage the recent achievements in parametric human models~\cite{loper2015smpl}.
Specifically, we align and \textcolor{black}{register} the generated 3D avatar to a SMPL mesh. So that it can be driven by the SMPL skeletons.

CLIP is only trained with static images and insensitive to sequential motions. Therefore, it is inherently challenging to generate reasonable motion sequences using only the supervision from CLIP. To tackle this problem, we divide the whole process into two stages: 1) generating candidate poses with the guidance of CLIP, and 2) synthesizing smooth and valid motions with the candidate poses as references. In the first stage, a code-book consisting of diverse poses is created by clustering. Poses that \textcolor{black}{match} motion descriptions are selected by CLIP from the code-book.
These generated poses serve as important clues for the second stage. A motion VAE is utilized in the second stage to learn motion priors, which facilitates the reference-guided motion synthesis.

With careful designs of the whole pipeline, \nickname{} is capable of generating high-quality 3D avatars with reasonable motion sequences guided by input texts as shown in Fig.~\ref{fig:teaser}. To evaluate our framework quantitatively, comprehensive user studies are conducted in terms of both avatar generation and animation to show our superiority over existing solutions. Moreover, qualitative experiments are also performed to validate the effectiveness of each component in our framework.

To sum up, our contributions are listed below: \textbf{1)} To the best of our knowledge, it is the first text-driven full avatar synthesis pipeline that includes the generation of shape, texture, and motion. \textbf{2)} Incorporating the power of large-scale pre-trained models, the proposed framework demonstrates strong zero-shot generation ability. Our avatar generation pipeline is capable of robustly generating animation-ready 3D avatars with high-quality texture and geometry. \textbf{3)} Benefiting from the motion VAE, a novel zero-shot text-guided reference-based motion synthesis approach is proposed. \textbf{4)} Extensive qualitative and quantitative experiments show that the generated avatars and motions are of higher quality compared to existing methods and are highly consistent with the corresponding input natural languages.

%% file: sections/02_related_work.tex
\section{Related Work}
\label{sec:relatedwork}

\paragraph{Avatar Modeling and Generation.}
For its wide application in industries, human modeling has been thoroughly studied for decades. Driven by a large-scale human body dataset~\cite{pishchulin17pr}, SMPL~\cite{loper2015smpl} and SMPL-X~\cite{pavlakos2019expressive} are proposed as a parametric human model.
For its strong interpretability and compatibility with the modern graphics pipeline, we choose SMPL as the template of our avatar. However, they only provide the ability to model the human body without clothes. Many efforts~\cite{bhatnagar2020combining, jiang2020bcnet, bhatnagar2019multi, hong2021garmentd} have been made on the modeling of clothed humans.

Due to the complexity of clothes and accessories, non-parametric human modelings~\cite{corona2021smplicit, palafox2021npms, saito2021scanimate, burov2021dynamic, mihajlovic2021leap, habermann2021real, huang2020arch, weng2019photo, alldieck2018video} are proposed to offer more flexibility on realistic human modeling.
Moreover, inspired by recent advances in volume rendering, non-parametric human modeling methods based on the neural radiance field (\ie NeRF~\cite{mildenhall2020nerf, jain2021putting}) have also been studied ~\cite{peng2021neural, zhao2021humannerf, liu2021neural, peng2021animatable, habermann2021real, xu2021h}.
Combining advantages of volume rendering and SDF, NeuS~\cite{wang2021neus} is proposed recently to achieve high-quality geometry and color reconstruction. That justifies our choice of NeuS as base representations of avatars.

With the rapid development of deep learning in recent years, impressive progresses have been shown on 2D image generation. 2D face generation~\cite{karras2019style, karras2020analyzing, jiang2021talk, fu2022styleganhuman} is now a very mature technology for the simplicity of the face structure and large-scale high-quality face datasets~\cite{liu2015faceattributes}.
Recent works~\cite{han2018viton, sarkar2021humangan, lewis2021tryongan, jiang2022text2human} have also demonstrated wonderful results in terms of 2D human body generation and manipulation. Without the knowledge of 3D space, it has always been a difficult task to animate the 2D human body~\cite{siarohin2019animating, siarohin2019first, yoon2021pose, sarkar2021style}.

The 3D human generation~\cite{chen2022gdna, noguchi2021neural, noguchi2022unsupervised} is barely explored until the very recent. Combining the powerful SMPL-X and StyleGAN, StylePeople~\cite{grigorev2021stylepeople} proposes a data-driven animatable 3D avatar generation method.
Inspired by the advancements in NeRF-GAN~\cite{chan2021pi}, a recent work~\cite{zhang20213d} proposes to bring in 3D awareness to traditional 2D generation pipelines.

\paragraph{Motion Synthesis.}
Serving as one of the most significant parts of animation, motion synthesis has been the research focus of many researchers.
Several large-scale datasets~\cite{ionescu2013human3, vonMarcard2018, mehta2017monocular, mahmood2019amass, varol17_surreal, cai2021playing, cai2022humman} \textcolor{black}{provide} human \textcolor{black}{motions} as sequences of 3D keypoints or SMPL parameters.
The rapid \textcolor{black}{advancements} of motion datasets stimulate researches on the motion synthesis. Early works apply classical machine learning to unconditional motion synthesis~\cite{rose1998verbs,ikemoto2009gen,mukai2005geo}. DVGANs~\cite{xiao2014human}, Text2Action~\cite{hyemin2018text2action} and Language2Pose~\cite{ahuja2019lang} generate motions conditioned on short texts using fully annotated data. Action2Motion~\cite{guo2020action2motion} and Actor~\cite{petrovich2021action} condition the motion generation on pre-selected action classes. These methods require large amounts of data\cite{hong2022versatile} with annotations of action classes or language descriptions, which limits their applications. On the contrary, our proposed \nickname{} can drive the human model by general natural languages without any paired data. Some other works~\cite{li2021ai,aggarwal2021dance2music} focus on music-conditioned motion synthesis. Moreover, some works~\cite{bergamin2019drecon,won2019learning} focus on the physics-based motion synthesis. They construct motion sequences with physical constrains for more realistic generation results.

\input{figures/pipeline}

\paragraph{Zero-shot Text-driven Generation.}
Text-to-image synthesis~\cite{mansimov2015generating} has long been studied.
The ability to zero-shot generalize to unseen categories is first \textcolor{black}{shown} by \cite{reed2016generative}.
CLIP and DALL-E~\cite{ramesh2021zero} further \textcolor{black}{show} the incredible text-to-image synthesis ability by excessively scale-up the size of training data. Benefiting from the zero-shot ability of CLIP, many amazing zero-shot text-driven applications~\cite{frans2021clipdraw, xu2021simple, patashnik2021styleclip} are being developed.
Combining CLIP with 3D representations like NeRF or mesh, zero-shot text-driven 3D object generation~\cite{michel2021text2mesh, jain2021zero, sanghi2021clip, jetchev2021clipmatrix} and manipulation~\cite{wang2021clip} have also come true in recent months.

%% file: figures/pipeline.tex
\begin{figure*}[ht]
    \centering
    \includegraphics[width=\linewidth]{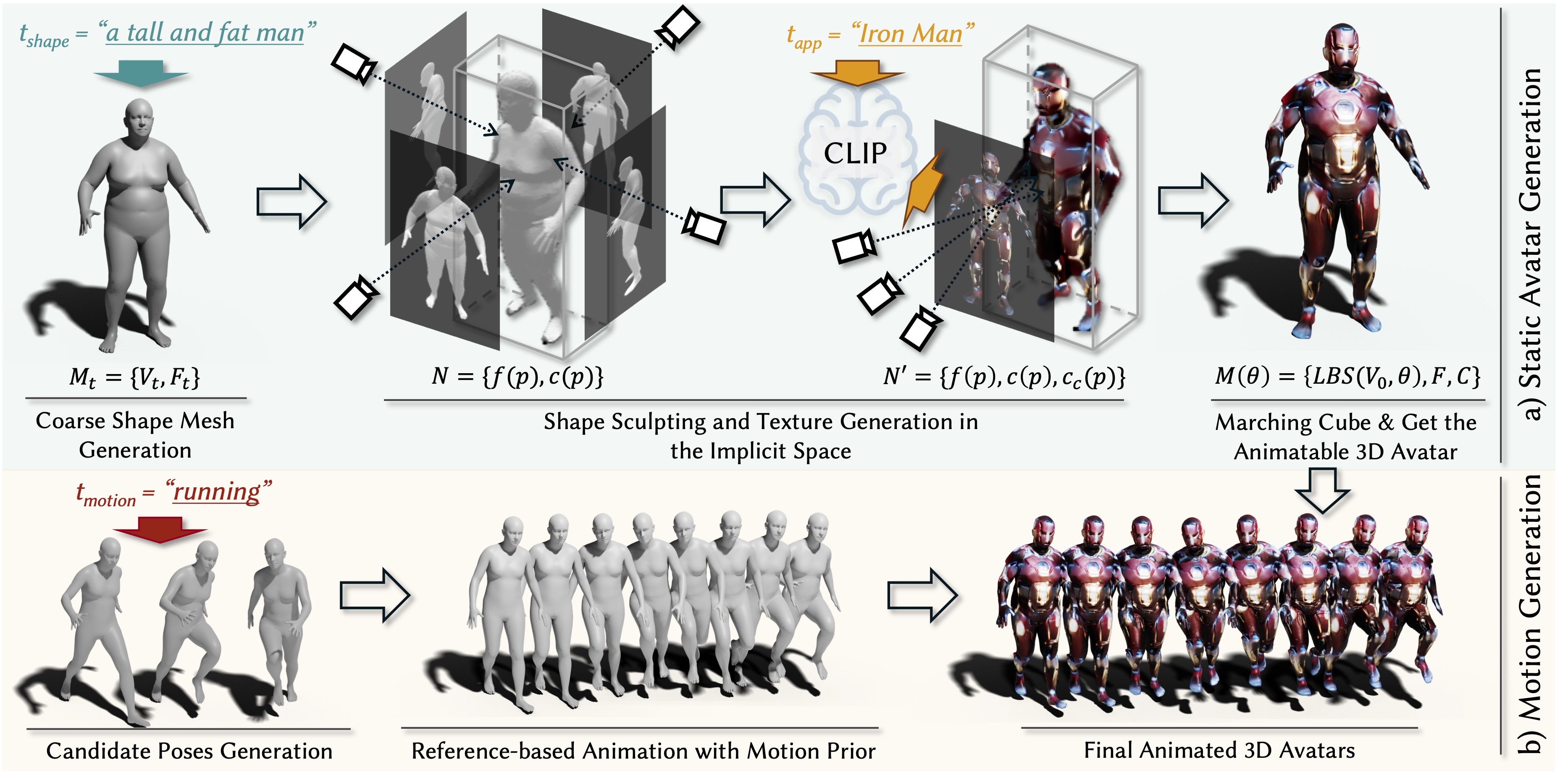}
    \caption{\textbf{An Overview of the Pipeline of \nickname{}.} The whole pipeline is divided into two parts: a) Static Avatar Generation; b) Motion Generation. Assume the user want to generate `a tall and fat Iron Man that is running'. An animatable avatar is generated guided by $t_{\textrm{shape}}$ = `a tall and fat man' and $t_{\textrm{app}}$ = `Iron Man'. Then a motion sequence matching the description $t_{\textrm{motion}}$ = `running' is generated to animate the generated avatar.}
    \Description{An Overview of the Pipeline of \nickname{}.}
    \label{fig:pipeline}
\end{figure*}

%% file: sections/03_methods.tex
\section{Our Approach} \label{sec:method}

Recall that our goal is zero-shot text-driven 3D avatar generation and animation, which can be formally defined as follows. The inputs are natural languages, $\textrm{text} = \{t_{\textrm{shape}}, t_{\textrm{app}}, t_{\textrm{motion}}\}$. The three texts \textcolor{black}{correspond} to the descriptions of the desired body shape, appearance and motion. The output is two-part, including \textbf{a)} an animatable 3D avatar represented as a mesh $M = \{V, F, C\}$, where $V$ is the vertices, $F$ stands for faces, $C$ represents the vertex colors; \textbf{b)} a sequence of poses $\Theta = \{\theta_i\}_{i=1}^{L}$ comprising the desired motion, where $L$ is the length of the sequence.

\subsection{Preliminaries} \label{sec:pre}

\paragraph{CLIP}~\cite{radford2021learning} is a vision-language pre-trained model trained with large-scale image-text datasets. It consists of an image encoder $E_I$ and a text encoder $E_T$. The encoders are trained in the way that the latent codes of paired images
and texts
are pulled together and unpaired ones are pushed apart. The resulting joint latent space of images and texts enables the zero-shot text-driven generation by encouraging the latent code of the generated content to align with the latent code of the text description. Formally, the CLIP-guided loss function is defined as
\begin{align}
    \mathcal{L}_{\textrm{clip}}(I, T) = 1 - \textrm{norm}(E_I(I)) \cdot \textrm{norm}(E_T(T)), \label{eq:clip}
\end{align}
where $(\cdot)$ represents the cosine distance. By minimizing $\mathcal{L}_{\textrm{clip}}(I, T)$, the generated image $I$ is encouraged to match the description of the text prompt $T$.

\paragraph{SMPL}~\cite{loper2015smpl} is a parametric human model driven by large-scale aligned human surface scans~\cite{pishchulin17pr}. The SMPL model can be formally defined as $M_{\textrm{SMPL}}(\beta, \theta; \Phi)$, where $\beta \in \mathbb{R}^{10}$ controls the body shapes, $\theta \in \mathbb{R}^{24\times 3}$ contains the axis angles of each joint.
In this work, we use SMPL as the template meshes for the initialization of the implicit 3D avatar and the automatic skeleton-binding and animation.

\paragraph{NeuS}~\cite{wang2021neus} proposes a novel volume rendering method that combines the advantages of SDF and NeRF~\cite{mildenhall2020nerf}, leading to high-quality surface reconstruction as well as photo-realistic novel view rendering. For some viewing point $o$ and viewing direction $v$, the color of the corresponding pixel is accumulated along the ray by
\begin{align}
    C(o, v) = \int_0^{\infty}w(t)c(p(t), v)dt, \label{eq:neus}
\end{align}
where $p(t) = o + vt$ is some point along the ray, $c(p(t), v)$ output the color at point $p(t)$, which is implemented by MLPs. $w(t)$ is a weighting function for the point $p(t)$. $w(t)$ is designed to be unbiased and occlusion-aware, so that the optimization on multi-view images can lead to the accurate learning of a SDF representation. It is defined as
\begin{align}
    w(t) = \frac{\phi_s(f(p(t)))}{\int_0^{\infty}\phi_s(f(p(u)))du}, \label{eq:wt}
\end{align}
where $\phi_s$ is the logistic density distribution, $f$ is an SDF network.
NeuS
is used in our work as the representation of the implicit 3D avatar.

\subsection{Pipeline Overview} \label{sec:pipeline}

As shown in Fig.~\ref{fig:pipeline}, the pipeline of our \nickname{} can be divided to two parts, \ie static avatar generation and motion generation. For the first part, the natural language description of the shape $t_{\textrm{shape}}$ is used for the generation of a coarse body shape $\beta_t$. Together with a pre-defined standing pose $\theta_{\textrm{stand}}$, we get the template mesh $M_t = \{V_t, F_{\textrm{SMPL}}\}$, where vertices $V_t = M(\beta_t, \theta_{\textrm{stand}}; \Phi)$ and faces $F_{\textrm{SMPL}}$ are given by SMPL. $M_t$ is then rendered to multi-view images for the training of a NeuS model $N$, which is later used as the initialization of our implicit 3D avatar. Guided by the appearance description $t_{\textrm{app}}$, $N$ is further optimized by CLIP in a shape-preserving way for shape sculpting and texture generation. After that, the target static 3D avatar mesh $M = \{V, F, C\}$ is extracted from $N'$ by the marching cube algorithm~\cite{lorensen1987marching} and aligned with $M_t$ to get ready for animation. For the second part, the natural language description of the motion $t_{\textrm{motion}}$ is used to generate candidate poses from a pre-calculated code-book. Then the candidate poses are used as references for the optimization of a pre-trained motion VAE to get the desired motion sequence.
\input{figures/coarseshape_genearation}

\subsection{Static Avatar Generation} \label{sec:avatar}

\subsubsection{Coarse Shape Generation} \label{sec:coarseshapegeneration}

As the first step of generating avatars, we propose to generate a coarse shape as the template mesh $M_t$ from the input description of the shape $t_{\textrm{shape}}$. For this step, SMPL is used as the source of possible human body shapes. As shown in Fig.~\ref{fig:coarse}, we first construct a code-book by clustering body shapes. Then CLIP is used to extract features of the renderings of body shapes in the code-book and texts to get the best matching body shape. Although the process is straightforward, careful designs are required to make full use of both powerful tools.

For the code-book construction, it is important to sample uniformly to cover most of the possible body shapes. The SMPL shape space $\beta \in \mathbb{R}^{10}$ is learned by the PCA decomposition, which results in non-uniform distribution. Therefore, we propose to train a shape VAE for a more uniform latent shape space. The code-book $B_{\textrm{shape}}\in\mathbb{R}^{K\times d_s}$ is constructed by K-Means clustering on the latent space of the shape VAE, which can be decoded to a set of body meshes $M_B = \{V_B, F_{SMPL}\}$, where $K$ is the size of the code-book, $d_s$ is the dimension of the latent code.

For the CLIP-guided code-book query, it is important to design reasonable query scores. We observe that for some attributes like body heights, it is hard to be determined only by looking at the renderings of the body without any reference. Therefore, we propose to set a reference and let CLIP score the body shapes in a relative way, which is inspired by the usage of CLIP in 2D image editing~\cite{patashnik2021styleclip}. We define a neutral state body shape $M_n$ and the corresponding neutral state text $t_n$ as the reference. The scoring of each entry $s_i$ in the code-book is defined as
\begin{align}
    s_i = 1 - \textrm{norm}(\Delta f_I) \cdot \textrm{norm}(\Delta f_T),
\end{align}
where $\Delta f_I = E_I(\mathcal{R}(M_B[i])) - E_I(\mathcal{R}(M_n))$, $\Delta f_T = E_T(t_{\textrm{shape}}) - E_T(t_n)$, $\mathcal{R}(\cdot)$ denotes a mesh renderer, $E_I$, $E_T$ represents CLIP image and text encoders. $\Delta f_T$ is the relative direction guided by the text, which is intended to be aligned with the visual relative shape differences $\Delta f_I$ in the CLIP latent space. Then the entry with the maximum score among the code-book is retrieved as the coarse shape $M_t = M_B[\textrm{argmax}_i(s_i)]$.

\input{figures/implicit_shape_init}
\input{figures/avatar_generation}

\subsubsection{Shape Sculpting and Texture Generation} \label{sec:shapetexturegeneration}

The generated template mesh $M_t$ represents a desired coarse naked body shape. To generate high-quality 3D avatars, the shape and texture need to be further sculpted and generated to match the description of the appearance $t_{\textrm{app}}$. As discussed previously, we choose to use an implicit representation, \ie NeuS, as the base 3D representation in this step for its advantages in both geometry and colors. In order to speed up the optimization and more importantly, control the general shape of the avatar for the convenience of animation, this step is designed as a two-stage optimization process.

The first stage creates an equivalent implicit representation of $M_t$ by optimizing a randomly initialized NeuS $N$ with the multi-view renderings $\{I_{M_t}^{i}\}$ of the template mesh $M_t$.
Specifically, as shown in Fig.~\ref{fig:init}, the NeuS $N = \{f(p), c(p)\}$ comprises of two sub-networks. The SDF network $f(p)$ takes some point $p$ as input and outputs the signed distance to its nearest surface. The color network $c(p)$ takes some point $p$ as input and outputs the color at that point. Both $f(p)$ and $c(p)$ are implemented using MLPs. It should be noted that compared to the original implementation of NeuS, we omit the color network's dependency on the viewing direction. The viewing direction dependency is originally designed to model the shading that changes with the viewing angle. In our case, it is redundant to model such shading effects since our goal is to generate 3D avatars with consistent textures, which ideally should be albedo maps. Similar to the original design, $N$ is optimized by a three-part loss function
\begin{align}
    \mathcal{L}_{1} = \mathcal{L}_{\textrm{color}} + \lambda_{1} \mathcal{L}_{\textrm{reg}} + \lambda_{2} \mathcal{L}_{\textrm{mask}}. \label{eq:l1}
\end{align}
$\mathcal{L}_{\textrm{color}}$ is the reconstruction loss between the \textcolor{black}{rendered} images and the ground truth multi-view \textcolor{black}{renderings} $\{I_{M_t}^{i}\}$. $\mathcal{L}_{\textrm{reg}}$ is an Eikonal term~\cite{gropp2020implicit} to regularize the SDF $f(p)$. $\mathcal{L}_{\textrm{mask}}$ is a mask loss that encourages the network to only reconstruct the foreground object, \ie the template mesh $M_t$.
The resulting NeuS $N$ serves as an initialization for the second stage optimization.

The second stage stylizes the initialized NeuS $N$ from the first stage by the power of CLIP. At the same time, the coarse shape $M_t$ should still be maintained, resulting in two possible solutions. The first one is to fix the weights of $f(p)$, which ensures the final \textcolor{black}{generated} shape is the same as $M_t$. The color network $c(p)$ is optimized to `colorize' the fixed shape. However, as discussed before, $M_t$ is generated from SMPL, which only provides the shape of a naked human body. Purely adding textures on the template shapes, although simple and tractable, is not considered the optimal solution. Because not only textures but also geometry is crucial in the process of avatar creation. To allow the fine-level shape sculpting, we need to fine-tune the weights of $f(p)$. At the same time, to maintain the general shape of the template mesh, $c(p)$ should maintain its original function of reconstructing the template mesh to allow the reconstruction loss during the optimization. This leads to the second solution, where we keep the original two sub-networks $f(p)$ and $c(p)$ and introduce an additional color network $c_c(p)$. Both color networks share the same SDF network $f(p)$ as shown in Fig.~\ref{fig:avatar_generation}. $f(p)$ and $c(p)$ are in charge of the reconstruction of the template mesh $M_t$. $f(p)$ together with $c_c(p)$ are responsible for the stylizing part and comprises the targeting implicit 3D avatar. Formally, the new NeuS model $N' = \{f(p), c(p), c_c(p)\}$ now consists of three sub-networks, where $f(p)$ and $c(p)$ are initialized by the pre-trained $N$, and $c_c(p)$ is a randomly initialized color network. All three sub-networks are trainable. Similarly, we ignore the viewing direction dependency in the additional color network.

The second-stage optimization is supervised by a three-part loss function 
\begin{align}
    \mathcal{L}_2 = \mathcal{L}_1 + \lambda_{3} \mathcal{L}_{\textrm{clip}}^{c} + \lambda_{4} \mathcal{L}_{\textrm{clip}}^{g},
\end{align}
where $\mathcal{L}_1$ is the reconstruction loss over NeuS $\{f(p), c(p)\}$ as defined in Eq.~\ref{eq:l1}. $\mathcal{L}_{\textrm{clip}}^{c}$ and $\mathcal{L}_{\textrm{clip}}^{g}$ are CLIP-guided loss functions that guide the texture and geometry generation to match the description $t_1$ using two different types of renderings, which are introduced as follows.

The first type of the rendering is the colored rendering $I_c$ of the NeuS model $\{f(p), c_c(p)\}$, which is calculated by applying Eq.~\ref{eq:neus} to each pixel of the image. The second type is the texture-less rendering $I_g$ of the same NeuS model. The rendering algorithm used here is ambient and diffuse shading. For each ray $\{o, v\}$, the normal direction $n(o, v)$ of the first surface point it intersects with can be calculated by the accumulation of the gradient of the SDF function at each position along the ray, which is formulated as
\begin{align}
    n(o, v) = \int_0^{\infty} w(t) \nabla f(p(t)) dt,
\end{align}
where $w(t)$ is the weighting function as defined in Eq.~\ref{eq:wt}. To render the texture-less model, a random light direction needs to be sampled. To avoid the condition where the light and camera are at opposite sides of the model and the geometry details of the model cannot be revealed, the light direction is uniformly sampled in a small range around the camera direction. Formally, defining the camera direction in the spherical coordinate system as polar and azimuthal angles $\{\theta_c, \phi_c\}$, then the light direction $l$ is sampled from $\{\theta_c + X_1, \phi_c + X_2\}$, where $X_1, X_2 \sim \mathcal{U}(-\pi/4, \pi/4)$. Since coloring is not needed for the texture-less rendering, we could simply calculate the gray level of the ray $\{o, v\}$ by
\begin{align}
    C_{gray}(o, v) = A + D \times n(o, v) \cdot l, \label{eq:gray}
\end{align}
where $A \sim \mathcal{U}(0, 0.2)$ is randomly sampled from a uniform distribution, $D = 1 - A$ is the diffusion. By applying Eq.~\ref{eq:gray} to each pixel of the image, we get the texture-less rendering $I_g$.
In practice, we find that random shading on textured renderings $I_c$ benefiting the uniformity of the generated textures, which is discussed later in the ablation study. The random shading on $I_c$ is similar to the texture-less rendering, which is formally defined as
\begin{align}
    C_{shade}(o, v) = A + D \times n(o, v) \cdot l * C(o, v).
\end{align}
Then the two types of CLIP-guided loss functions are formally defined as $\mathcal{L}_{\textrm{clip}}^{c} = \mathcal{L}_{\textrm{clip}}(I_c, t_{\textrm{app}})$, $\mathcal{L}_{\textrm{clip}}^{g} = \mathcal{L}_{\textrm{clip}}(I_g, t_{\textrm{app}})$.
%

To render a $H\times W$ image, a total of $H \times W \times Q$ queries need to be performed given $Q$ query times for each ray. Due to the high memory footprint of volume rendering, the rendering resolution $H \times W$ is heavily constrained. In our experiment on one 32GB GPU, the maximum $H_{\textrm{max}}$ and $W_{\textrm{max}}$ are around $110$, which is far from the resolution upper bound of $224$ provided by CLIP. With the network structures and other hyper-parameters not modified, to increase the rendering resolution, we propose a dilated silhouettes-based rendering strategy based on the fact that the rays not encountering any surface do not contribute to the final rendering while consuming large amounts of memories. Specifically, we could get the rays that have high chances of encountering surfaces by calculating the silhouette of the rendering of the template mesh $M_t$ with the given camera parameter. Moreover, to ensure a proper amount of shape sculpting space is allowed, we further dilate the silhouettes. The rays within the silhouettes are calculated and make contributions to the final rendering. Defining the ratio between the area of the dilated silhouette and the total area of the image as $r_s$, this rendering strategy dynamically increases the maximum resolution to $H_{\textrm{max}}' \times W_{\textrm{max}}' = H_{\textrm{max}} \times W_{\textrm{max}} / r_s$. Empirically, the resolution can be increased to around $150^2$.

In order to further increase the robustness of the optimization process, three augmentation strategies are proposed: \textbf{a)} random background augmentation; \textbf{b)} random camera parameter sampling; \textbf{c)} semantic-aware prompt augmentation. They are introduced as follows.

\input{figures/randombgaug}

Inspired by Dream Fields~\cite{jain2021zero}, random background augmentation helps CLIP to focus more on the foreground object and prevents the generation of randomly floating volumes. As shown in Fig.~\ref{fig:randombgaug}, we randomly augment the backgrounds of the renderings to 1) pure black background; 2) pure white background; 3) Gaussian noises; 4) Gaussian blurred chessboard with random block sizes.

To prevent the network from finding short-cut solutions that only give reasonable renderings for several fixed camera positions, we randomly sample the camera extrinsic parameters for each optimization iteration in a manually-defined importance sampling way to get more coherent and smooth results. We choose to work with the `look at' mode of the camera, which consists of a look-at point, a camera position, and an up direction. We set the up direction to always align with the up direction of the template body. The look at position is sampled from a Gaussian distribution $X, Y, Z \sim \mathcal{N}(0, 0.1)$, which are then clipped between $-0.3$ and $0.3$ to prevent the avatar from being out of the frame. The camera position is sampled using a spherical coordinate system with the radius $R$ sampled from a uniform distribution $\mathcal{U}(1, 2)$, the polar angle $\theta_c$ sampled from a uniform distribution $\mathcal{U}(0, 2\pi)$, the azimuthal angle $\phi_c$ sampled from a Gaussian distribution $\mathcal{N}(0, \pi/3)$ for the camera to point at the front side of the avatar in the most of the iterations.

\input{figures/promptaug}

So far, no human prior is introduced in the whole optimization process, which may result in generating textures at wrong body parts or not generating textures for important body parts, which will later be shown in the ablation study.
To prevent such problems, we propose to explicitly bring in human prior in the optimization process by augmenting the prompts in a semantic-aware manner. For example, as shown in Fig.~\ref{fig:promptaug}, if $t_{\textrm{app}}=$`Steve Jobs', we would augment $t_{\textrm{app}}$ to two additional prompts $t_{\textrm{face}}=$`the face of Steve Jobs' and $t_{\textrm{back}}=$`the back of Steve Jobs'. For every four iterations, the look-at point of the camera is set to be the center of the face to get renderings of the face, which will be supervised by the prompt $t_\textrm{face}$. The `face augmentation' directly supervises the generation of the face, which is important to the quality of the generated avatar since humans are more sensitive to faces. For the second augmented prompt, when the randomly sampled camera points at the back of the avatar, $t_\textrm{back}$ is used as the corresponding text to explicitly guide the generation of the back of the avatar.

\subsubsection{Make the Static Avatar Animatable}

With the generated implicit 3D avatar $N'$, the marching cube algorithm is performed to extract its meshes $M=\{V, F, C\}$ before making it animatable. Naturally, with the careful design of keeping the overall shape unchanged during the optimization, the generated mesh can be aligned with the initial template mesh. Firstly, the nearest neighbor for each vertex in $V$ is retrieved in the vertices of the template mesh $M_t$. Blend weights of each vertex in $V$ are copied from the nearest vertex in $M_t$. Secondly, an inverse LBS algorithm is used to bring the avatar's standing pose $\theta_{\textrm{stand}}$ back to the zero pose $\theta_0$. The vertices $V$ are transformed to $V_{\theta_0}$. Finally, $V_{\theta_0}$ can be driven by any \textcolor{black}{pose} $\theta$ using the LBS algorithm. Hence, for any pose, the animated avatar can be formally defined as $M(\theta) = (\textrm{LBS}(V_{\theta_0}, \theta), F, C)$.

\subsection{Motion Generation} \label{sec:motion}

Empirically, CLIP is not capable of directly estimating the similarities between motion sequences and natural language descriptions. It also lacks the ability to assess the smoothness or rationality of motion sequences. These two limitations suggest that it is hard to generate motions only using CLIP supervision. Hence, we have to introduce other modules to provide motion priors. However, CLIP has the ability to value the similarity between a rendered human pose and a description. Furthermore, similar poses can be regarded as references for the expected motions. Based on the above observations, we propose a two-stage motion generation process: 1) candidate poses generation guided by CLIP. 2) motion sequences generation using motion priors with candidate poses as references. The details are illustrated as follows.

\input{figures/candidate_poses}

\subsubsection{Candidate Poses Generation} \label{sec:candidateposegeneration}

To generate poses consistent with the given description $t_{\textrm{motion}}$, an intuitive method is to directly optimize the parameter $\theta$ in the SMPL model or the latent code of a pre-trained pose VAE (\eg VPoser~\cite{pavlakos2019expressive}).
However, they can hardly yield reasonable poses due to difficulties during optimization, which are later shown in the experiments. Hence, it is not a wise choice to directly optimize the poses.

As shown in Fig.~\ref{fig:candidate}, to avoid the direct optimization, we first create a code-book from AMASS dataset~\cite{mahmood2019amass}.
To reduce dimensions, VPoser is used to encode poses to $z \in \mathbb{R}^{d_p}$.
Then we use K-Means to acquire $K$ cluster centroids that form our pose code-book $B_{z} =\mathbb{R}^{K \times d_p}$. Each element of $B_{z}$ is then decoded by VPoser, which leads to a set of poses $B_\theta$.

Given the motion description $t_{\textrm{motion}}$, we calculate the similarity between $t_{\textrm{motion}}$ and each pose $B_{\theta}[i]$ from the code-book $B_{\theta}$, which can be defined as
\begin{align}
    s_i & = 1 - \text{norm}(E_I(\mathcal{R}(\theta_B[i]))) \cdot \text{norm}(E_T(t_{motion})).
\end{align}
Top-k scores $s_i$ and their corresponding poses $B_\theta[i]$ are selected to construct the candidate pose set $S$, which \textcolor{black}{serves} as references to generate a motion sequence in the next stage.

\input{figures/motion_VAE}

\input{figures/animation}

\input{figures/overall_results}

\subsubsection{Reference-Based Animation with Motion Prior} \label{sec:referencebasedanimation}

We propose a two-fold method to generate a target motion sequence that matches the motion description $t_\textrm{motion}$. 1) A motion VAE is trained to capture human motion priors. 2) We optimize the latent code of the motion VAE using candidate poses $S$ as references. The details are introduced as follows.

\paragraph{Motion VAE Pre-train.}
We take inspirations from Actor~\cite{petrovich2021action} to construct the motion VAE.
Specifically, as shown in Fig.~\ref{fig:mvae}, the motion VAE contains three parts: the motion encoder $E_{\textrm{motion}}$, the reparameterization module, and the motion decoder $D_{\textrm{motion}}$.
A motion sequence is denoted as $\Theta^{+} \in \mathbb{R}^{L \times 24 \times 6}$, where $L$ represents the length of the motion. Each joint is represented by a continuous 6-D tensor~\cite{zhou2019continuity}.

The motion encoder $E_{\textrm{motion}}$ contains a projection layer, a positional embedding layer, several transformer encoder layers and an output layer. Formally, $E_{\textrm{motion}}$ can be defined as
$y = E_{\textrm{motion}}(\Theta^{\prime}) = E_o(\operatorname{TransEncoder}(\phi(E_p(\Theta^{\prime}))))$,
where $E_p$ and $E_o$ are fully connected layers,
$\phi$ represents positional embedding operation~\cite{vaswani2017attention}. $\operatorname{TransEncoder}$ includes multiple transformer encoder layers~\cite{vaswani2017attention}.

The reparameterization module yields a Gaussian distribution where $\mu=R_{\mu}(y),\sigma=R_{\sigma}(y)$. $R_{\mu}, R_{\sigma}$ are fully connected layers to calculate the mean $\mu$ and the standard deviation $\sigma$ of the distribution, respectively. Using the reparameterization trick~\cite{kingma2013auto}, a random latent code $z_{\textrm{motion}}$ is sampled under the distribution $\mathcal{N}(\mu, \sigma)$. The latent code is further decoded by $D_{\textrm{motion}}$, which can be formally defined as
$\Theta^{\ast} = D_\textrm{motion}(z_\textrm{motion}) = D_o(\operatorname{TransDecoder}(D_p(z_{\textrm{motion}}), \phi(\mathbf{0})))$,
where $D_p$ and $D_o$ are fully connected layers,
$\mathbf{0}$ is a zero-vector. $\operatorname{TransDecoder}$ contains several transformer decoder layers.
A loss function with two terms is proposed to train the motion VAE, which is defined as
\begin{align}
    \mathcal{L}_{\textrm{mVAE}} &=\lambda_{5} \cdot \mathcal{L}_{\textrm{KL}} + \mathcal{L}_{\textrm{recon}},
    \label{eq:mVAE}
\end{align}
where $\lambda_{5}$ is a hyper-parameter to balance two terms. $\mathcal{L}_{\textrm{KL}}$ computes the KL-divergence to enforce the distribution assumption. $\mathcal{L}_{\textrm{recon}}$ is the mean squared error between $\Theta^{+}$ and $\Theta^{*}$ for the reconstruction of the motion sequences.

\paragraph{Optimization on the Motion VAE.}
With the pre-trained motion VAE, we attempt to optimize its latent code $z_t$ to synthesize \textcolor{black}{a} motion sequence $\Theta=D_{\textrm{motion}}(z_t)$. As shown in Fig.~\ref{fig:animation}, three optimization constraints are proposed as
\begin{align}
    \mathcal{L}_{\textrm{motion}}=\mathcal{L}_{\textrm{pose}}+\lambda_6 \mathcal{L}_{\textrm{delta}}+\lambda_7 \mathcal{L}_{\textrm{clip}}^m,
    \label{eq:animation_loss}
\end{align}
where $\lambda_6,\lambda_7$ are hyper-parameters to balance the terms. $\mathcal{L}_{\textrm{pose}}$ is the reconstruction term between the decoded motion sequence $\Theta$ and candidate poses $S$.
$\mathcal{L}_{\textrm{delta}}$ measures the range of the motion to prevent the motion from being overly-smoothed.
$\mathcal{L}_{\textrm{clip}}^m$ encourages each single pose in the motion to match the input motion description.
Details of three loss terms are introduced as follows.

Given reference poses $S=\{\theta_1, \theta_2,\dots,\theta_k \}$, the target is to construct a motion sequence that is close enough to these poses. We propose to minimize the distance between $\theta_i$ and its nearest frame $\Theta_j$, where $i \in \{1, 2, \dots, k\}$, and $j \in \{1, 2, \dots, L\}$.
Note that we assume $\theta_i$ is less similar to $t_{\textrm{motion}}$ with larger $i$. Therefore, we use a coefficient $\lambda_{\textrm{pose}}(i)=1-\frac{i-1}{k}$ to focus more on candidate poses with higher similarities. Formally, the reconstruction loss is defined as
\begin{align}
    \mathcal{L}_{\textrm{pose}}=\sum\limits_{i=1}^k \lambda_{\textrm{pose}}(i) \min\limits_{j} \{\|\theta_i - \Theta_j\| \}.
\end{align}

Empirically, only using the reconstruction loss $\mathcal{L}_{\textrm{pose}}$, the generated motion tends to be over-smoothed. To generate motions with larger motion ranges, we design a motion range term $\mathcal{L}_{\textrm{delta}}$ to measure the smoothness of adjacent poses,
\begin{align}
    \mathcal{L}_{\textrm{delta}}= -\sum\limits_{i=1}^{L-1} \|\Theta_i - \Theta_{i+1}\|,
\end{align}
which serves as a penalty term against over-smoothed motions. More intense motions will be generated when increasing $\lambda_6$.

The matching scheme of the reconstruction term $\mathcal{L}_{pose}$ does not guarantee the ordering of candidate poses.
Lack of supervision on the pose orderings would lead to unstable generation results. Furthermore, candidate poses might only contribute to a small part of the final motion sequence, which would lead to unexpected motion pieces. To tackle these two problems, we design
an additional CLIP-guided loss term
\begin{align}
    \mathcal{L}_{\textrm{clip}}^m= \sum\limits_{i=1}^L \lambda_{\textrm{clip}}(i) \cdot s_i,
\end{align}
where $s_i$ is the similarity score between the pose $\Theta_i$ and text description $t_{\textrm{motion}}$, which is defined as
\begin{align}
    s_i = 1 - \operatorname{norm}(E_I(\mathcal{R}(\Theta_i))) \cdot \operatorname{norm}(E_T(t_{\textrm{motion}})).
\end{align}
$\lambda_{\textrm{clip}}(i)=\frac{i}{L}$ is a monotonically increasing function so that $\mathcal{L}_{\textrm{clip}}^m$ gives higher penalty to the later poses in the sequence. With the above CLIP-guided term, the whole motion sequence will be more consistent with $t_{\textrm{motion}}$. Empirically, we find that we only need to sample a small part of poses in $\Theta$ for the calculation of this term, which will speed up the optimization without observable degradation in performance.

%% file: figures/coarseshape_genearation.tex
\begin{figure}[t]
    \centering
    \includegraphics[width=\linewidth]{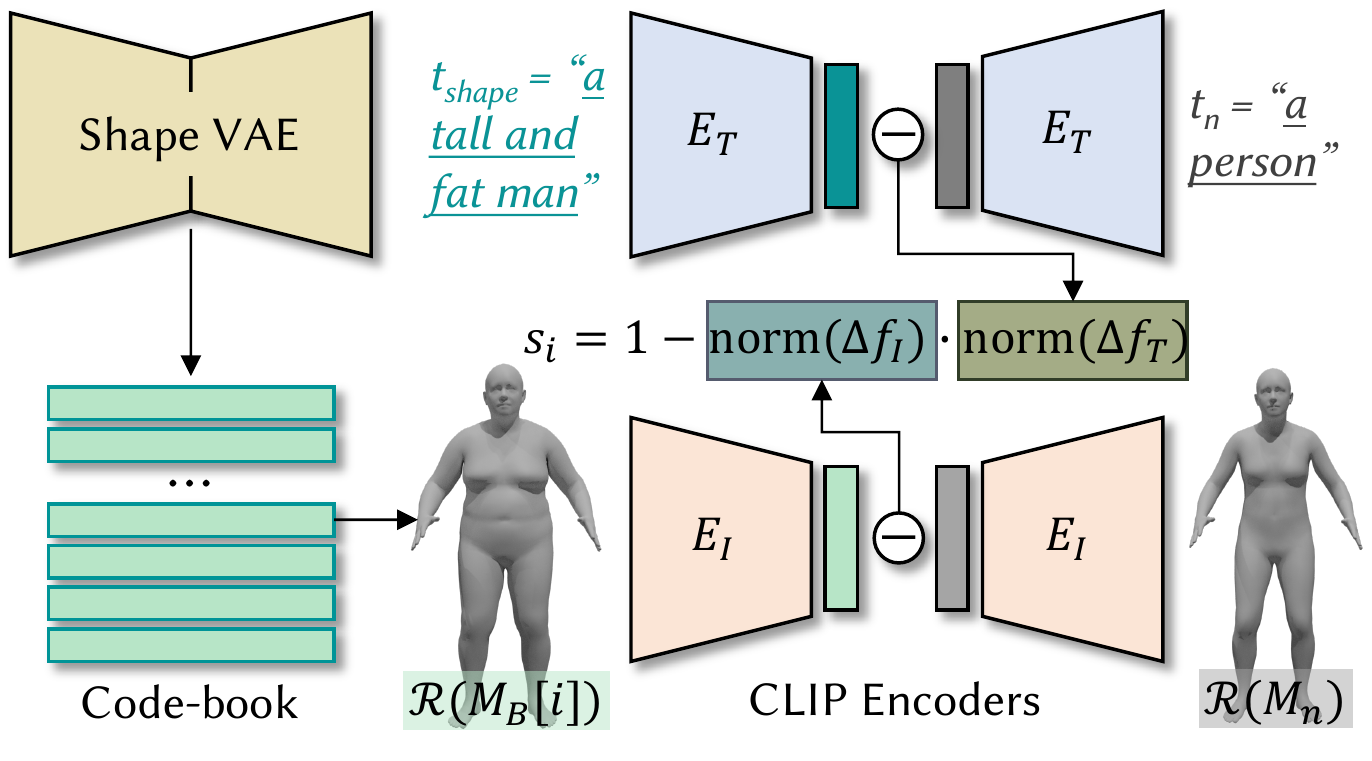}
    \caption{\textbf{Illustration of the Coarse Shape Generation.} A shape VAE is trained to construct a code-book which is used for CLIP-guided query to get a best match for the input text $t_\textrm{shape}$. To introduce the awareness of body attributes like height, a neutral shape $M_n$ and text $t_n$ is defined as the anchor. The relative direction in latent space is used for the CLIP-guided query.}
    \Description{Illustration of the Coarse Shape Generation.}
    \label{fig:coarse}
\end{figure}

%% file: figures/implicit_shape_init.tex
\begin{figure}[t]
    \centering
    \includegraphics[width=\linewidth]{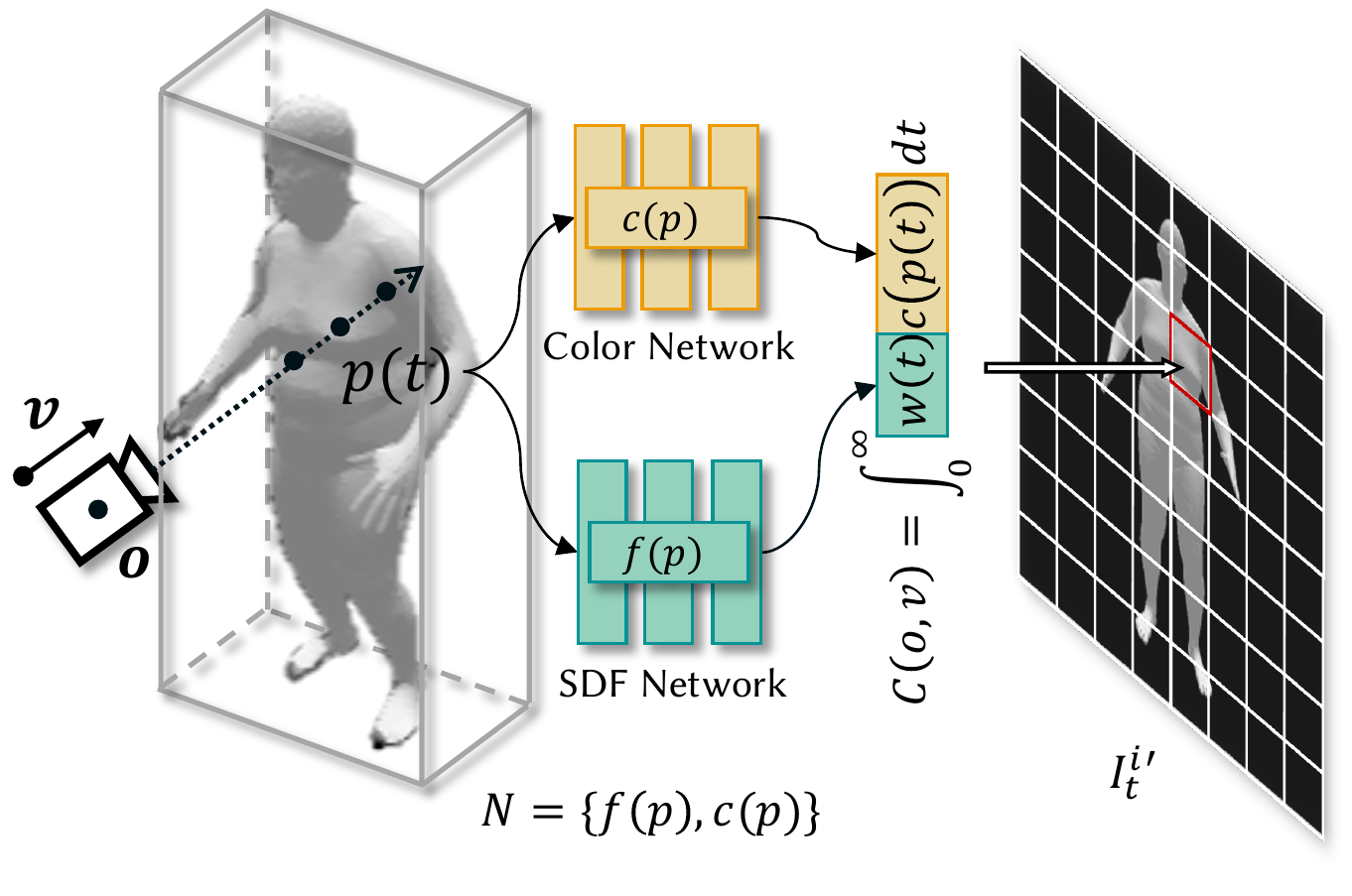}
    \caption{\textbf{Initialization of the Implicit 3D Avatar.} Multi-view renderings of the template mesh $M_t$ is used to optimize a randomly initialized NeuS network $N$, which is later used as an initialization of the implicit 3D avatar.}
    \Description{Initialization of the implicit 3D avatar.}
    \label{fig:init}
\end{figure}

%% file: figures/avatar_generation.tex
\begin{figure*}[ht]
    \centering
    \includegraphics[width=\linewidth]{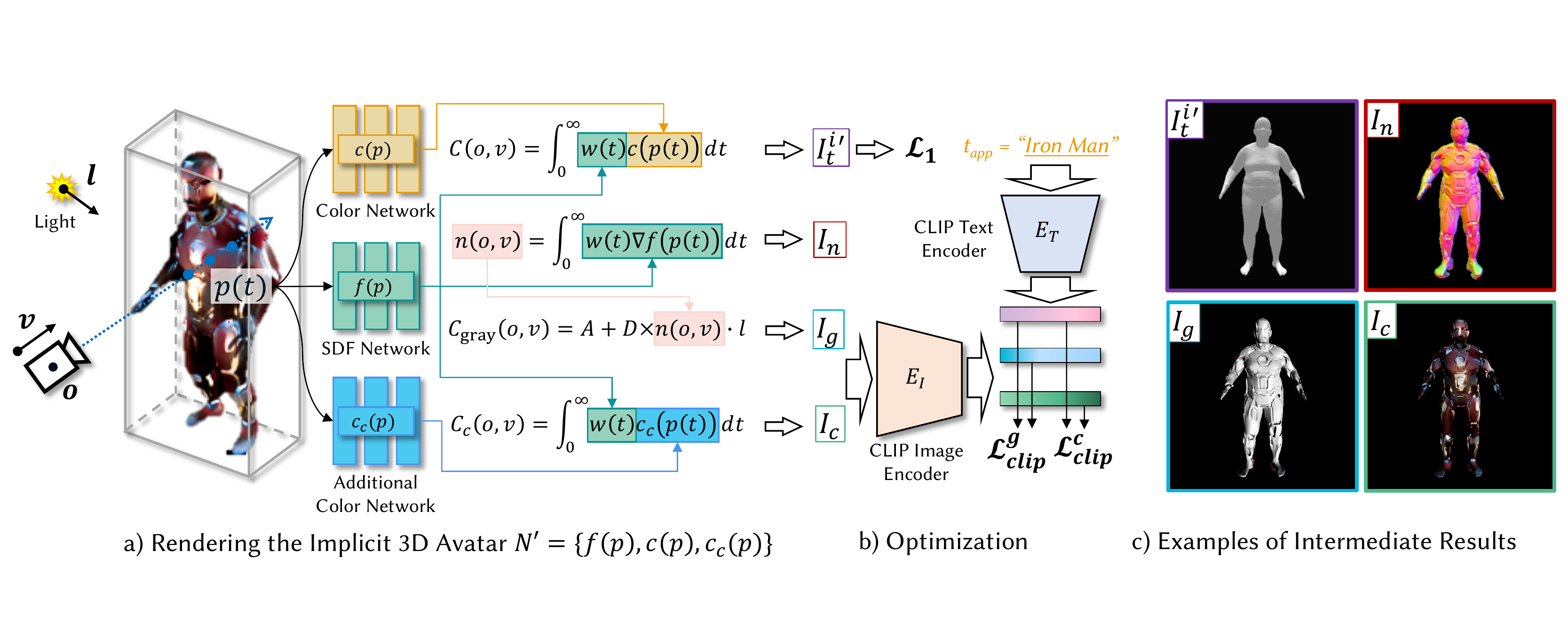}
    \caption{\textbf{Detailed Method of Shape Sculpting and Texture Generation.} An additional color network $c_c(p)$ is appended to the initialized implicit 3D avatar for texture generation, which is illustrated in a). Three types of constraints introduced for the optimization are shown in b), including reconstruction loss $\mathcal{L}_1$, CLIP-guided losses $\mathcal{L}^{g_{clip}}$ and $\mathcal{L}^{c}_{clip}$ for the geometry sculpting and texture generation, respectively. The sub-figure c) shows examples of intermediate results.}
    \Description{Detailed Method of Shape Sculpting and Texture Generation.}
    \label{fig:avatar_generation}
\end{figure*}

%% file: figures/randombgaug.tex
\begin{figure}[t]
    \centering
    \includegraphics[width=0.9\linewidth]{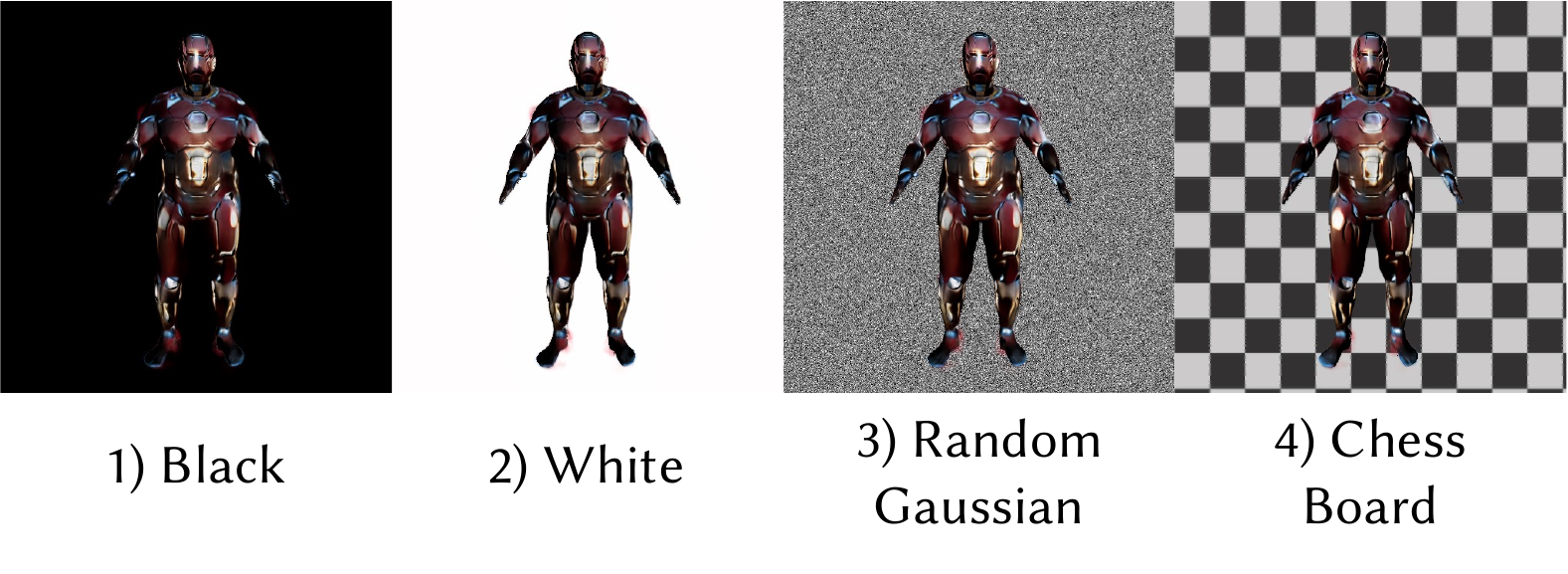}
    \caption{Examples of four types of random background augmentations.}
    \Description{Examples of the Random Background Augmentation.}
    \label{fig:randombgaug}
\end{figure}

%% file: figures/promptaug.tex
\begin{figure}[t]
    \centering
    \includegraphics[width=0.9\linewidth]{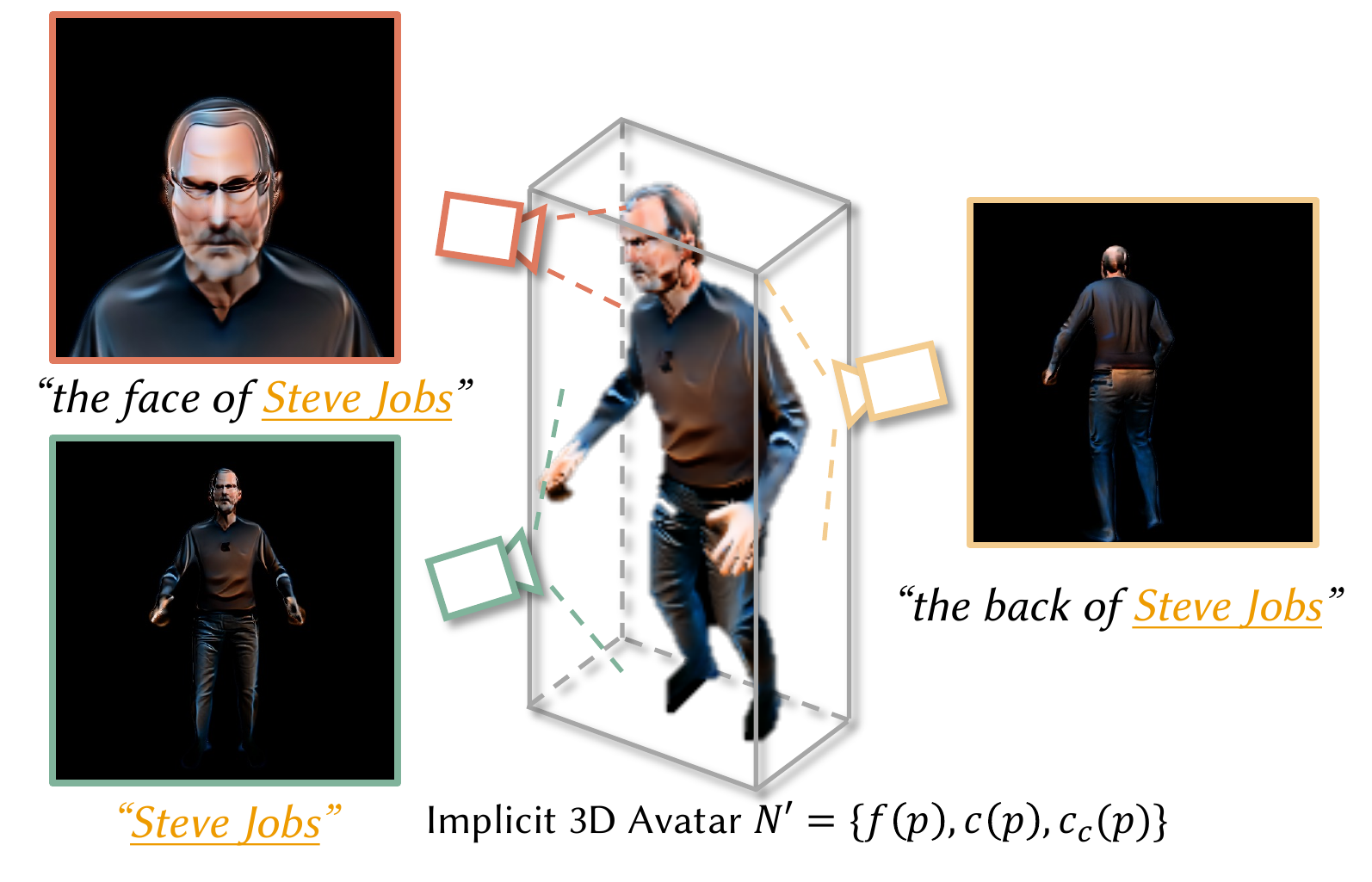}
    \caption{Examples of two types of prompt augmentations. One for the detailed \textcolor{black}{refinement} of the face and the other for the back side of the avatar.}
    \Description{Examples of the Prompt Augmentation.}
    \label{fig:promptaug}
\end{figure}

%% file: figures/candidate_poses.tex
\begin{figure}[t]
    \centering
    \includegraphics[width=0.9\linewidth]{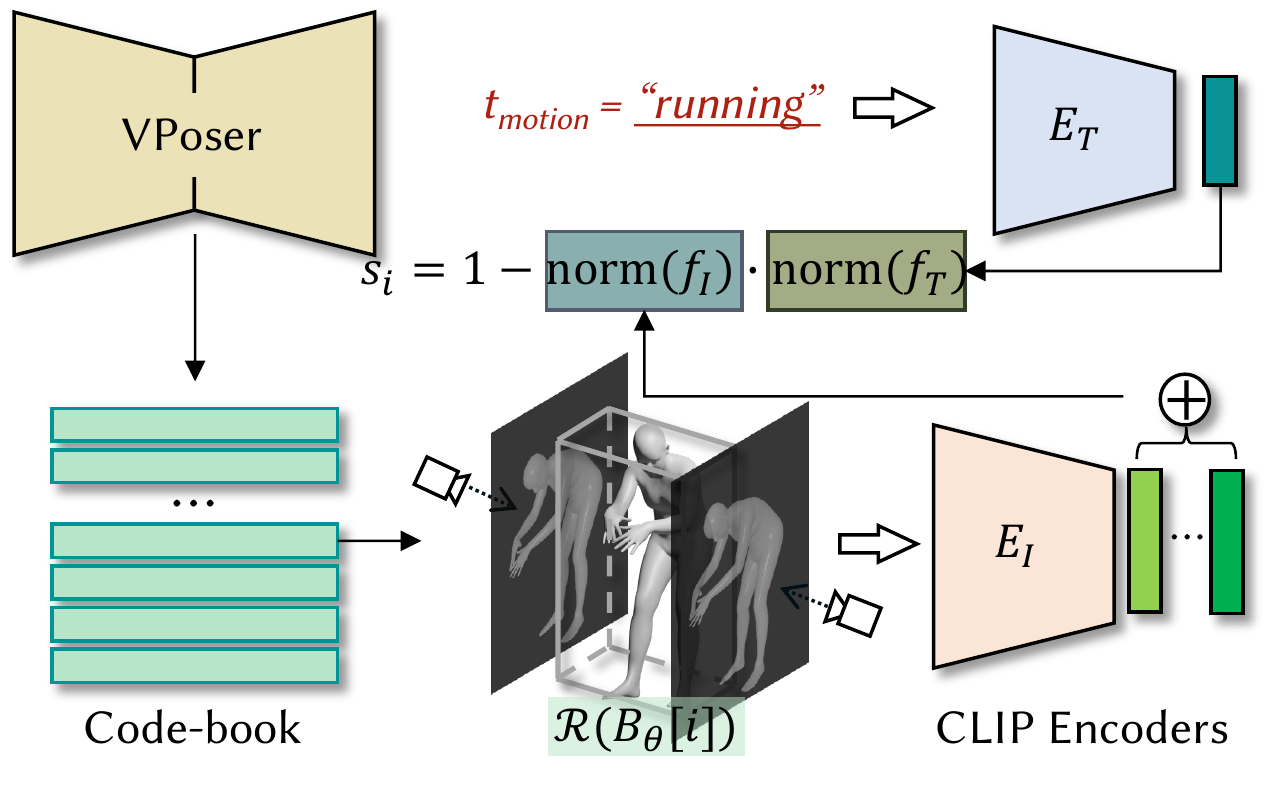}
    \caption{\textbf{Detailed Pipeline of Candidate Poses Generation.} The pre-trained VPoser is first used to build a code-book. Given text description $t_{\textrm{motion}}$, 
    each pose feature $f_I$ from the code-book is used to calculate the similarity with the text feature $f_T$, which is used to select Top-K entries as candidate poses.}
    \Description{Detailed Pipeline of Selecting Candidate Poses.}
    \label{fig:candidate}
\end{figure}

%% file: figures/motion_VAE.tex
\begin{figure}[t]
    \centering
    \includegraphics[width=0.9\linewidth]{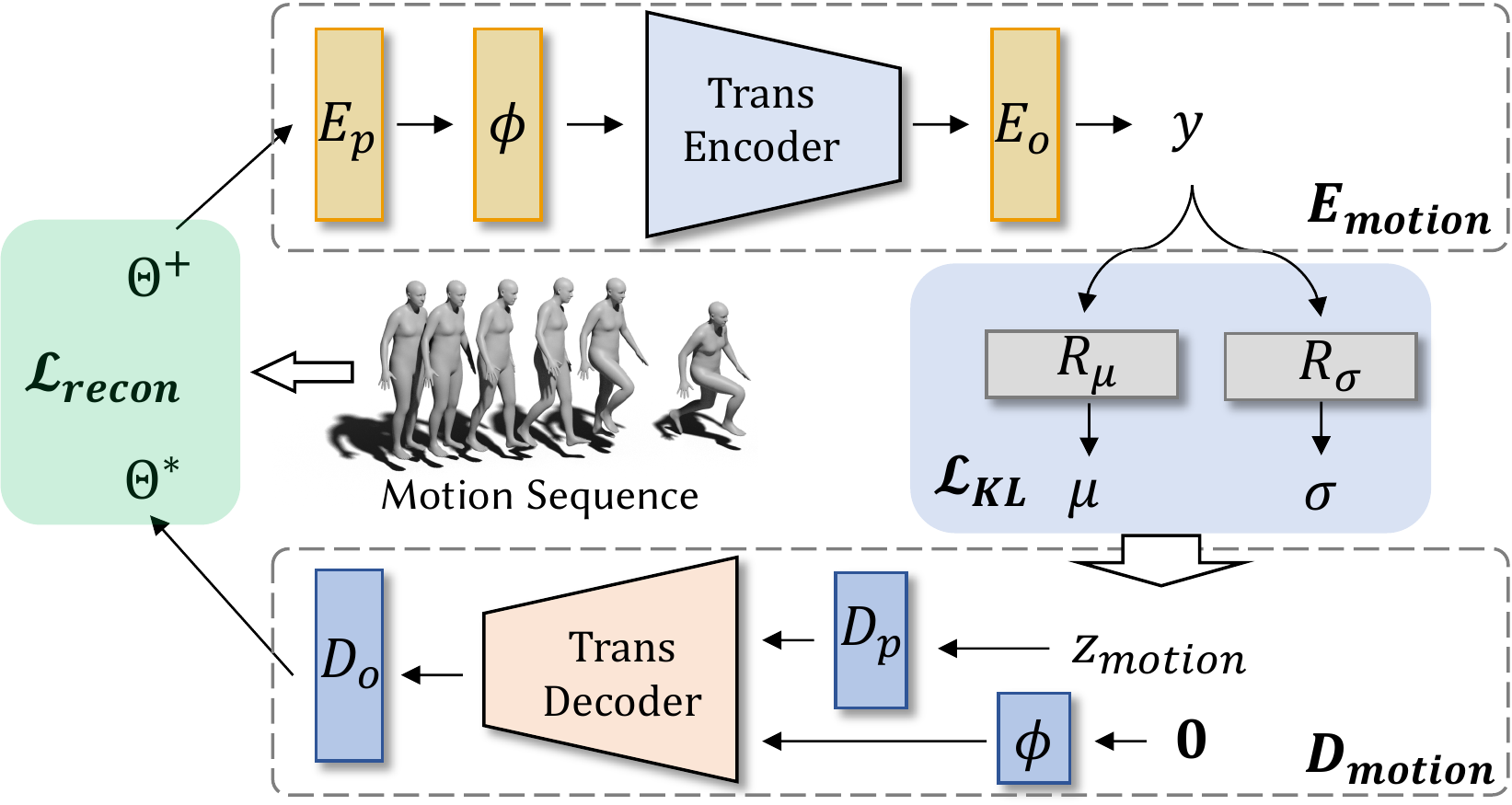}
    \caption{\textbf{Structure of the Motion VAE.} The motion VAE contains three parts: the encoder $E_{\textrm{motion}}$, the decoder $D_{\textrm{motion}}$, and a reparameterization module. 
    The reconstruction loss $\mathcal{L}_{\textrm{recon}}$ and the KL-divergence term $\mathcal{L}_{\textrm{KL}}$ are used for the motion VAE training.
    }
    \Description{Structure of motion VAE.}
    \label{fig:mvae}
\end{figure}

%% file: figures/animation.tex
\begin{figure*}[ht]
    \centering
    \includegraphics[width=0.9\linewidth]{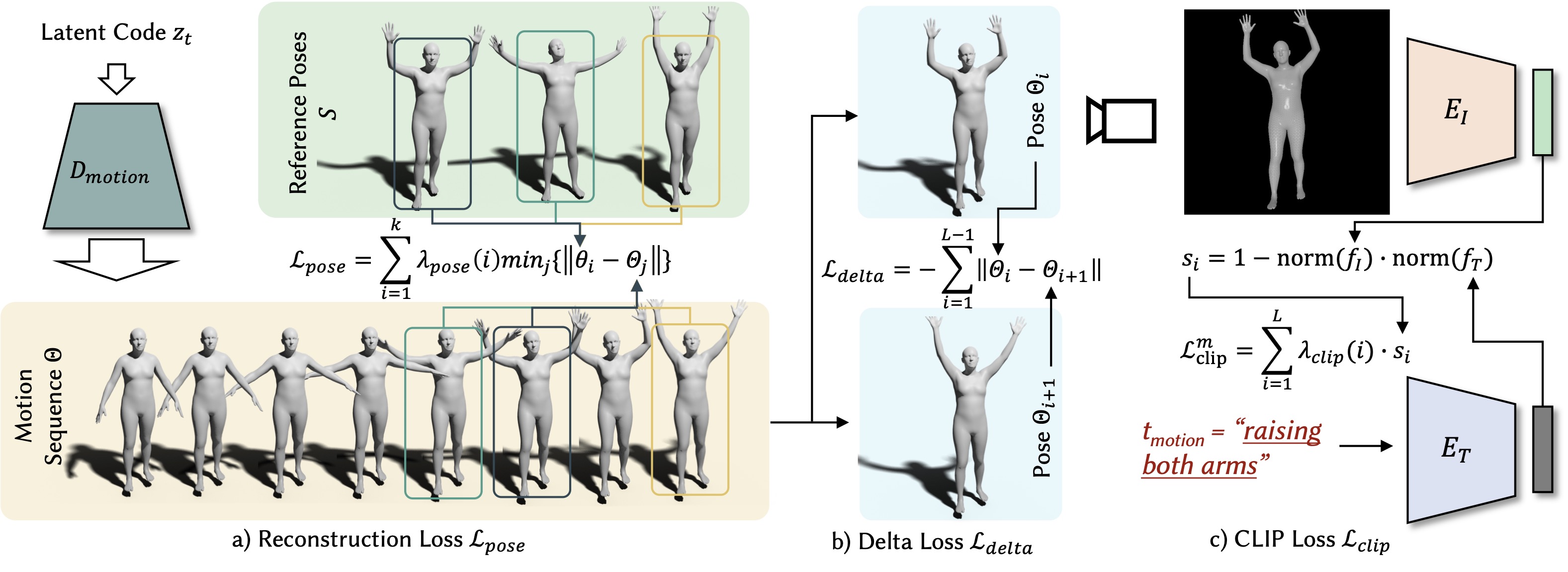}
    \caption{\textbf{Detailed Pipeline of Reference-Based Animation with Motion Prior.} Three constraint terms are designed to optimize the latent code $z_t$. For the motion sequence $\Theta$ decoded by $D_{\textrm{motion}}$, $\mathcal{L}_{\textrm{pose}}$ is used to minimize the distance between each candidate pose and the nearest pose in $\Theta$.
    $\mathcal{L}_{\textrm{delta}}$ is an adjustable loss item that measures the differences between adjacent poses and is capable of controlling the intensity of motion. 
    $\mathcal{L}_{\textrm{clip}}^{m}$ measures the similarity between description $t_{\textrm{motion}}$ and each pose in $\Theta$.}
    \Description{Detailed Pipeline of Reference-Based Motion Generation.}
    \label{fig:animation}
\end{figure*}

%% file: figures/overall_results.tex
\begin{figure*}[ht]
    \centering
    \includegraphics[width=\linewidth]{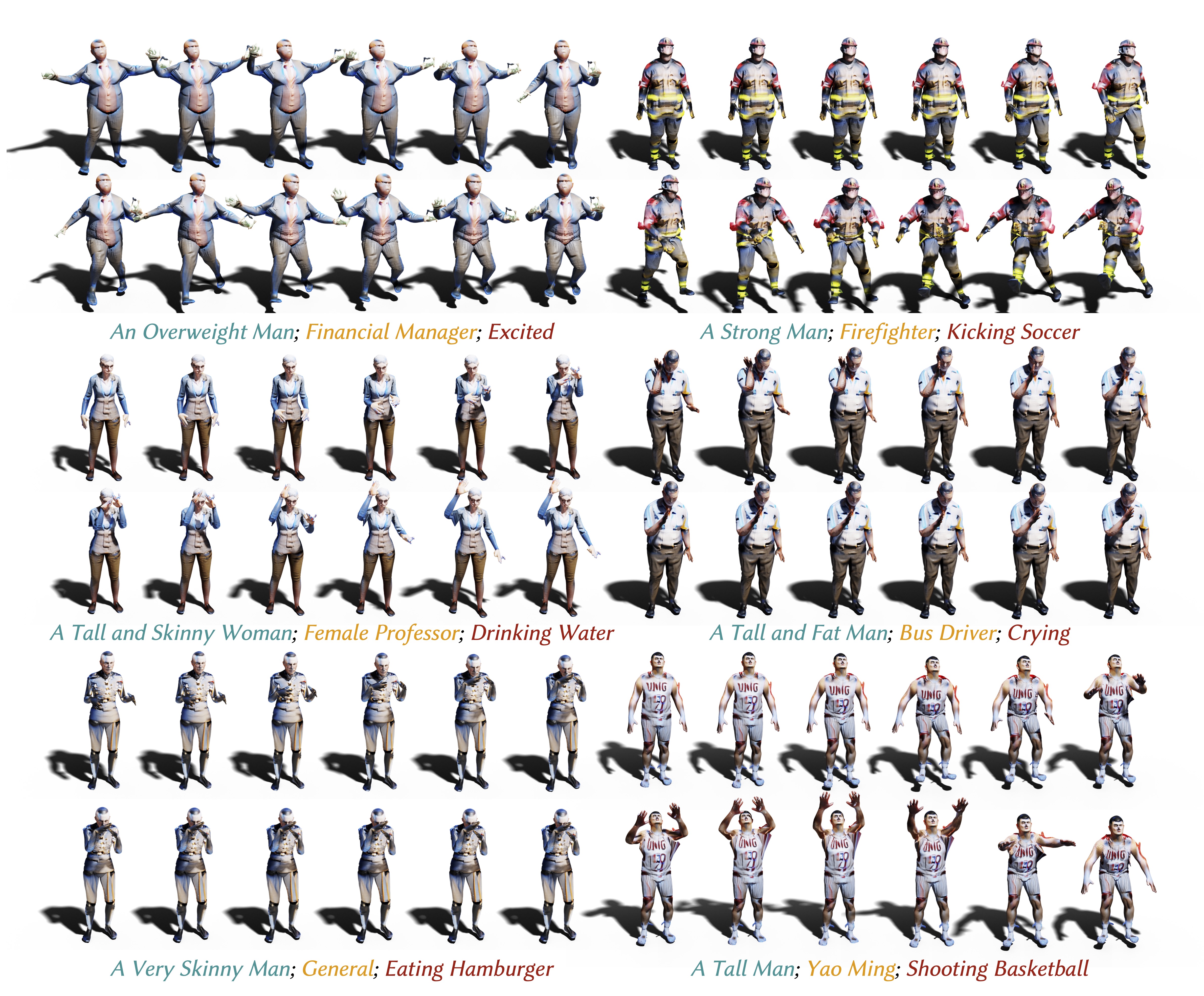}
    \caption{\textbf{Overall Results of \nickname{}.} Renderings of several animated 3D avatars are shown in sequence. The corresponding driving texts for \textcolor{shape}{shape}, \textcolor{appearance}{appearance} and \textcolor{motion}{motion} are put below the sequences.}
    \Description{Overall Results of \nickname{}.}
    \label{fig:overall_results}
\end{figure*}

%% file: sections/04_experiments.tex
\input{figures/avatar_ablation}

\section{Experiments}
\label{sec:exp}

\subsection{Experimental Setup}

\subsubsection{Implementation Details}
The shape VAE (Sec.~\ref{sec:coarseshapegeneration}) uses a two-layer MLP with a 16 dimension latent space for the encoder and decoder. Then 40,960 latent codes are sampled from a Gaussian distribution and clustered to 2,048 centroids by K-Means which composite the code-book. For the shape sculpting and texture generation (Sec.~\ref{sec:shapetexturegeneration}), we adjust the original NeuS such that the SDF network uses a 6-layer MLP and the color network uses a 4-layer MLP. For each ray, we perform 32 uniform samplings and 32 importance \textcolor{black}{samplings}. The Adam algorithm~\cite{kingma2014adam} with a learning rate of $5\times 10^{-4}$ is used for 30,000 iterations of optimization.

For candidate pose generation (Sec.~\ref{sec:candidateposegeneration}), we directly use the pre-trained VPoser~\cite{pavlakos2019expressive} and use K-Means to acquire 4,096 cluster centroids from the AMASS dataset~\cite{mahmood2019amass}. We select top-5 poses for the next stage (Sec.\ref{sec:referencebasedanimation}). Our proposed Motion VAE has a 256 dimension latent space.
The length of the motion is 60. We train 100 epochs for motion VAE on AMASS dataset. An Adam optimizer with a $5\times 10^{-4}$ learning rate is used.
During optimizing latent code for $t_{\textrm{motion}}$, Adam is used for 5,000 iterations of optimization with a $1\times10^{-2}$ learning rate.

\input{figures/coarseshape_all_results}

\subsubsection{Baselines}
\label{exp:baselines}
Though it is the first work to generate and animate 3D avatars in a zero-shot text-driven manner, we design reasonable baseline methods for the evaluation of each part of \nickname{}. For the coarse shape generation, we design a baseline method where the shape parameters (\ie the SMPL $\beta$ and latent code of the shape VAE) are directly optimized by CLIP-guided losses. For the shape sculpting and texture generation, we compare our design with Text2Mesh~\cite{michel2021text2mesh}. Moreover, we introduce a NeRF-based baseline by adapting Dream Fields~\cite{jain2021zero}, where our `additional color network' design is added to its pipeline to constraint the general shape of the avatar. 

For the candidate pose generation, three baseline methods are designed to compare with our method. To illustrate the difficulty of direct optimization, we set two baselines that directly optimize on SMPL parameter $\theta$ and latent code $z$ in VPoser. Moreover, inspired by CLIP-Forge~\cite{sanghi2021clip}, we use Real NVP~\cite{dinh2016density} to get a bi-projection between the normal distribution and latent space distribution of VPoser.
This normalization flow network is conditioned on the CLIP features.
This method does not need paired pose-text data.
As for the second part of the motion generation, we design two baseline methods to compare with: Baseline (i) first sorts candidate poses $S$ by their similarity scores $s_i$. Then, direct interpolations over the latent codes between each pair of adjacent poses are performed to generate the motion sequence. Baseline (ii) uses the motion VAE to introduce motion priors into the generative pipeline. But (ii) directly calculates the reconstruction loss without using the re-weighting technique.

\subsection{Overall Results}
Overall results of the whole pipeline of \nickname{} are shown in Fig.~\ref{fig:overall_results}.
Avatars with diverse body shapes along with varied appearances are generated with high quality. They are driven by generated motion sequences that are reasonable and consistent with the input descriptions. In a zero-shot style, \nickname{} is capable of generating animatable avatars and motions, making use of the strong prior in pre-trained models.
The whole process of avatar generation and animation, which originally requires expert knowledge of professional software, can now be simply driven by natural languages with the help of our proposed \nickname{}.

\subsection{Experiments on Avatar Generation}

\subsubsection{Ablation Study}
To validate the effectiveness of various designs in the avatar generation module, we perform extensive ablation studies. We ablate the designs of 1) background augmentation; 2) supervision on texture-less renderings; 3) random shading on the textured renderings; 4) semantic-aware prompt augmentation. The ablation settings shown in Fig.~\ref{fig:avatarablation} are formed by subsequently adding the above four designs to a baseline method where only textured renderings are supervised by CLIP.
As shown in the first two columns of Fig.~\ref{fig:avatarablation}, background augmentation has a great influence on the texture generation, without which the textures tend to be very dark. Comparing the second and third columns, adding the supervision on texture-less renderings improves the geometry quality by a large margin. The geometry of `Ablation 2' has lots of random bumps, which make the surfaces noisy. While the geometry of `Ablation 3' is smooth and has detailed wrinkles of garments.
As shown by the `Ablation 3' and `Ablation 4', adding random shadings on textured renderings helps the generation of more uniform textures. For example, the `Donald Trump' of `Ablation 3' has a brighter upper body than the lower one, which is improved in `Ablation 4'. Without the awareness of human body semantics, the previous four settings cannot generate correct faces for the avatars. The last column, which uses the semantic-aware prompt augmentation, has the best results in terms of the face generation.

\subsubsection{Qualitative Results of Coarse Shape Generation.}
For this part, we design two intuitive baseline methods where direct CLIP supervision is back-propagated to the shape parameters.
As shown in Fig.~\ref{fig:coarseshapeall} (a), both optimization methods fail to generate body shapes consistent with description texts. Even opposite text guidance (\eg `skinny' and `overweight') leads to the same optimization direction. In comparison, our method robustly generates reasonable body shapes agreeing with input texts. More diverse qualitative results of our method are shown in Fig.~\ref{fig:coarseshapeall} (b).
\input{figures/concept_mixing}

\subsubsection{Qualitative Results of Shape Sculpting and Texture Generation.}
\paragraph{Broad Range of Driving Texts.}
Throughout extensive experiments, our method is capable of generating avatars from a wide range of appearances descriptions including three types: 1) celebrities, 2) fictional characters and 3) general words that describe people, as shown in Fig.~\ref{fig:avatar_results}.
As shown in Fig.~\ref{fig:avatar_results} (a), given celebrity names as the appearance description, the most iconic outfit of the celebrity is generated. 
Thanks to our design of semantic-aware prompt augmentation, the faces are also generated correctly.
For the fictional character generation as illustrated in Fig.~\ref{fig:avatar_results} (b), avatars of most text descriptions can be correctly generated. Interestingly, for the characters that have accessories with complex geometry (\eg helmets of `Batman', the dress of `Elsa'), the optimization process has the tendency of `growing' new structures out of the template human body.
As for the general descriptions, our method can handle very broad ranges including common job names (\eg `Professor', `Doctor'), words that describe people at a certain age (\eg `Teenager', `Senior Citizen') and other fantasy professions (\eg `Witch', `Wizard'). It can be observed that other than the avatars themselves, their respective iconic \textcolor{black}{objects} can also be generated. For example, the `Gardener' grasps flowers and grasses in his hands.

\paragraph{Zero-Shot Controlling Ability.}
Other than the overall descriptions of the appearance as shown above, our method is also capable of zero-shot controlling at a more detailed level. As shown in Fig.~\ref{fig:moreconstrol}, we can control the faces generated on the avatars, \eg Bill Gates wearing an Iron Man suit, by tuning the semantic-aware prompt augmentation.
Moreover, we can control the clothing of the avatar by direct text guidance, \eg `Steve Jobs in white shirt'.

\paragraph{Concept Mixing.}
Inspired by DALL-E~\cite{ramesh2021zero} and Dream Fields~\cite{jain2021zero}, one of the most exciting applications of CLIP-driven generation is concept mixing. As shown in Fig.~\ref{fig:conceptmixing}, we demonstrate examples of mixing fantasy elements with celebrities. While maintaining the recognizable identities, the fantasy elements blend in with the generated avatars naturally.

\paragraph{Geometry Quality.}
The critical design of texture-less rendering supervision mainly \textcolor{black}{contributes} to the geometry generation. We mainly compare our \nickname{} with the adapted Dream Field, which is based on NeRF. As shown in Fig.~\ref{fig:geometrycompare}, our method consistently outperforms Dream Fields in terms of geometry quality. Detailed muscle shapes, armor curves, and cloth wrinkles can be generated.

\paragraph{Robustness.}
Other than the generation quality, we also investigate the robustness of our algorithm compared with the baseline method Text2Mesh. For each method, we use the same five random seeds for five independent runs for the same prompt.
As shown in Fig.~\ref{fig:robustness}, our method manages to output results with high quality and consistency with the input text. Text2Mesh fails most runs, which shows that the representations of meshes are unstable for optimization, especially with weak supervision.

\paragraph{Failure Cases.}
Although \textcolor{black}{a wide range of} generation results are experimented with and demonstrated above, there \textcolor{black}{exist} failure cases in generating loose garments and accessories. For example, as shown in Fig.~\ref{fig:avatar_results}, the dress of `Elsa' and the cloak of `Doctor Strange' are not generated. The \textcolor{black}{geometry} of exaggerated hair and beard (\eg the breaded Forrest Gump) \textcolor{black}{is} also challenging to \textcolor{black}{be} generated correctly. This is caused by the reconstruction loss in the optimization process. The `growing' is discouraged and very limited changes of the geometry are allowed.

\subsubsection{Quantitative Results}
\textcolor{black}{
To quantitatively evaluate the results of our avatar generation method, we ask $22$ volunteers to perform a user study in terms of 1) the consistency with input texts, 2) texture quality, and 3) geometry quality. We randomly select $8$ input texts that describe appearances. They are used for avatar generation by three methods, \ie Dream Field, Text2Mesh and our \nickname{}. For each sample, the volunteers are asked to score the results of three methods from $1$ to $5$ in terms of the above three aspects.
As illustrated in Fig.~\ref{fig:avataruserstudy}, our method consistently outperforms the other two baseline methods in all three aspects. Moreover, the standard deviations of our method are the lowest among the three methods, which also demonstrates the stable quality of our method.
}

\subsection{Experiments on Motion Generation}

\subsubsection{Qualitative Results of Motion Generation.} \label{sec:exp_motion}
\paragraph{Ablation Study.}
To evaluate the effectiveness of our design choices in the reference-based animation module (\ie the proposed three constraint terms, the usage of motion VAE as a motion prior), we compare our method with two baseline methods as shown in Fig.~\ref{fig:motion_ablation}.
For `Brush Teeth', (i) generates unordered and unrelated pose sequences.
(ii) also fails to generate reasonable motions, which is caused by its blind focus on reconstructing the unordered candidate poses.
By introducing a re-weighting mechanism, our method not only focuses on reconstruction but also considers the rationality of the generated motion. 
For `Kick Soccer', the motion sequences from (i) (ii) have no drastic changes when the leg kicks out. $\mathcal{L}_{\textrm{delta}}$ plays a significant role here to control the intensity of motions. As for `Raise Both Arms', it is supposed to generate a motion sequence from a neutral pose to a pose with raised arms. However, (i) generates a motion sequence that is contrary to the expected result. (ii) introduces several unrelated actions. With the help of $\mathcal{L}_{\textrm{clip}}^{m}$, our method is capable of generating motions with the correct ordering and better consistency with the descriptions.

\paragraph{Failure Cases.}
We also demonstrate some failure cases in Fig.~\ref{fig:motion_failure_case}. Throughout experiments, we find it hard to precisely control the body parts. For example, we cannot specifically control left or right hands, as shown in `raising left arm' case. Moreover, limited by the diversity of candidate poses, it is challenging to generate more complex motions like `hugging', `playing piano' and `dribbling'.

\input{figures/avatar_user_study}

\subsubsection{Qualitative Results of Candidate Pose Generation} \label{sec:exp_pose}
\paragraph{Comparison with Baseline Methods.}
Comparisons between different candidate pose generation methods are shown in Fig.~\ref{fig:poseresults} (a).
Both direct optimization methods (i) (ii) fail to generate reasonable poses, let alone poses that are consistent with the given description.
These results suggest that direct optimization on human poses parameters is intractable.
Compared with (i) and (ii), conditioned Real NVP (iii) can yield rational poses. However, compared to the proposed solution based on the code-book, the generated poses from (iii) are less reasonable and of lower quality.

\paragraph{Broad Range of Driving Texts.}
To demonstrate the zero-shot ability of the pose generation method, we experiment with four categories of motion descriptions: 1) abstract emotion descriptions (\eg `tired' and `sad'); 2) common action descriptions (\eg `walking' and `squatting'); 3) descriptions of motions related to body parts (\eg `raising both arms', `washing hands') 4) descriptions of motions that \textcolor{black}{involve} interaction with objects (\eg `shooting basketball'). They are shown in Fig.\ref{fig:poseresults} (b).

\subsubsection{Quantitative Results}
Quantitatively, we evaluate the candidate pose generation and reference-based animation separately. For the candidate pose generation, we ask $58$ volunteers to select the candidate poses that are the most consistent with the given text inputs for $15$ randomly selected samples. The percentage of the selected times of each method for each text input is used as the scores for counting. As shown in Fig.~\ref{fig:motionuserstudy} (a), compared with the baseline methods introduced in Sec.~\ref{sec:exp_pose}, our method outperforms them by a large margin.
\textcolor{black}{For the reference-based animation, we ask $20$ volunteers to score the results from 1 to 5 in terms of the consistency with the input texts and the overall quality for $10$ randomly selected samples. As shown in Fig.~\ref{fig:motionuserstudy} (b), compared with the two baseline methods introduced in Sec.~\ref{sec:exp_motion}, our method outperforms them by large margins in both consistency and quality.
}

\input{figures/motion_user_study}

\input{figures/avatar_results}
\input{figures/morecontrol}
\input{figures/geometrycompare}
\input{figures/robustness}
\input{figures/motion_ablation}
\input{figures/pose_results}
\input{figures/motion_failure_cases}

%% file: figures/avatar_ablation.tex
\begin{figure*}[ht]
    \centering
    \includegraphics[width=\linewidth]{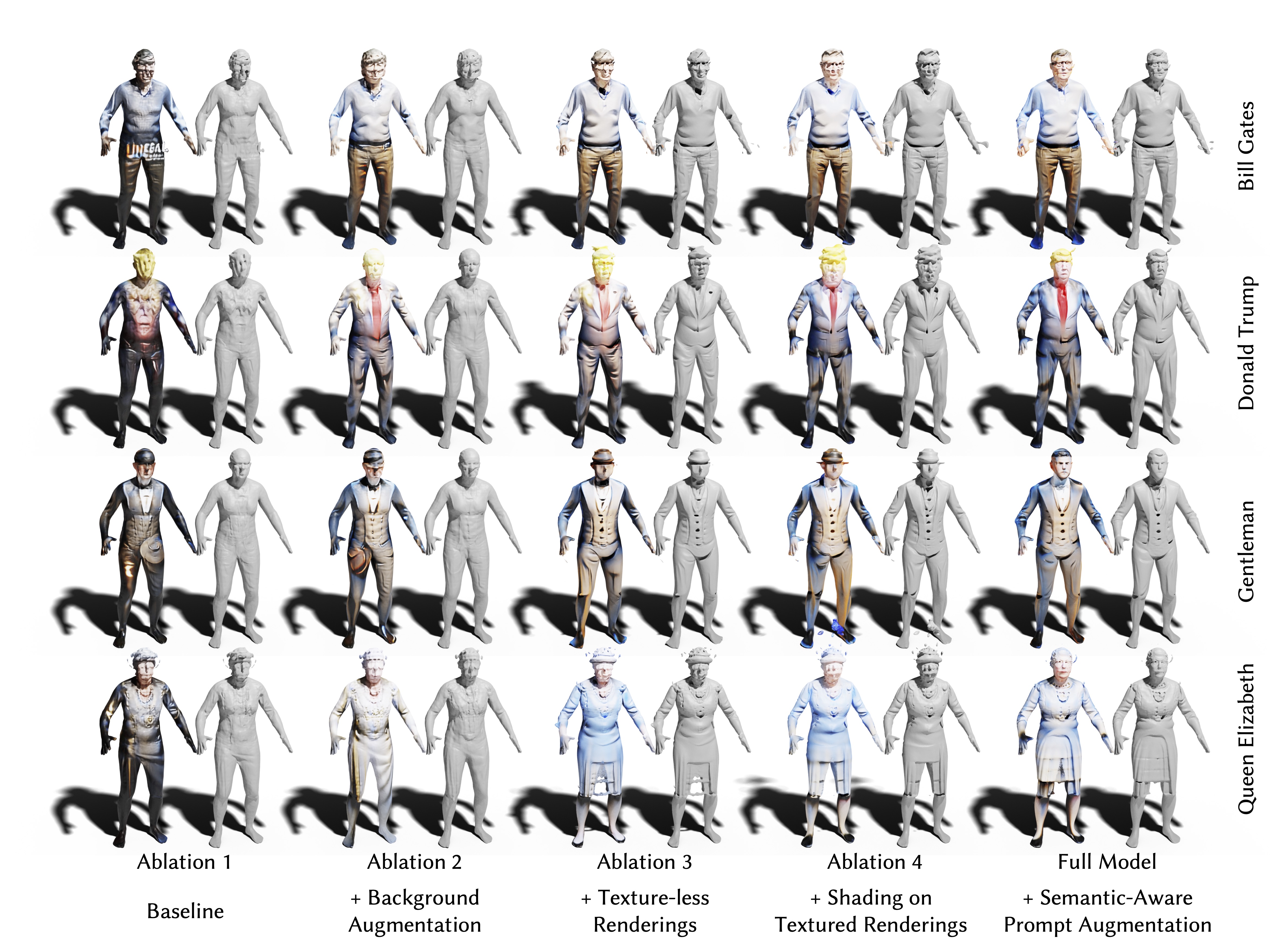}
    \caption{\textbf{Ablation Study on Static Avatar Generation.} Four ablation studies are performed to validate our design choices in avatar generation. Specifically, four ablation settings subsequently add 1) background augmentation, 2) texture-less renderings, 3) shading on textured renderings, 4) semantic-aware prompt augmentation.}
    \Description{Ablation Study on Static Avatar Generation.}
    \label{fig:avatarablation}
\end{figure*}

%% file: figures/coarseshape_all_results.tex
\begin{figure*}[ht]
    \centering
    \includegraphics[width=\linewidth]{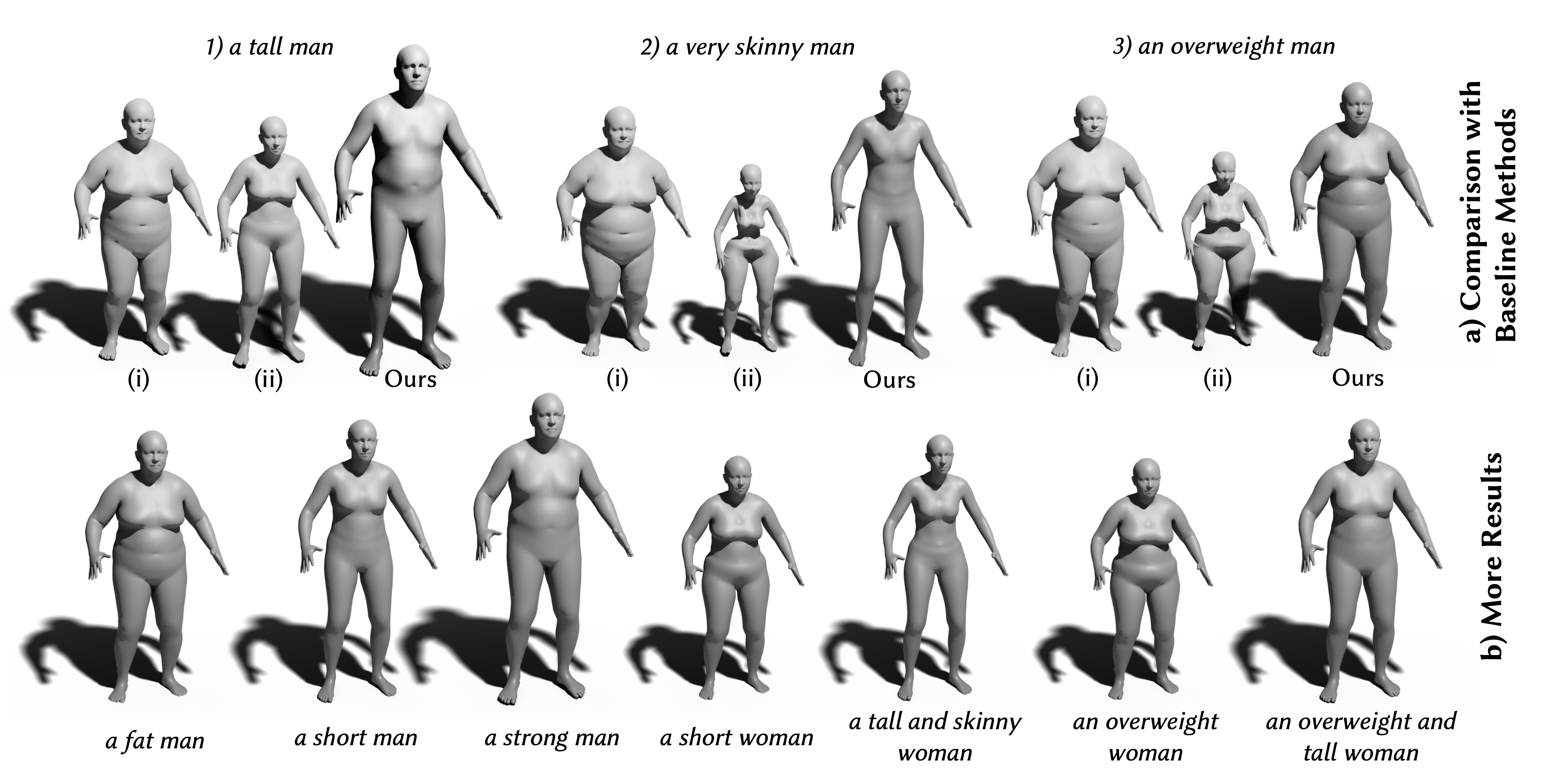}
    \caption{\textbf{Results of the Coarse Shape Generation.} a) The qualitative comparison between our method and two baselines (i) direct optimization over SMPL $\beta$ space, (ii) direct optimization over VAE latent space. Two baseline methods fail to generate reasonable body shapes. While our method successfully \textcolor{black}{generates} body shapes that match the descriptions above. b) More results of the coarse shape generation are shown.
    }
    \Description{Results of the Coarse Shape Generation.}
    \label{fig:coarseshapeall}
\end{figure*}

%% file: figures/concept_mixing.tex
\begin{figure*}[ht]
    \centering
    \includegraphics[width=\linewidth]{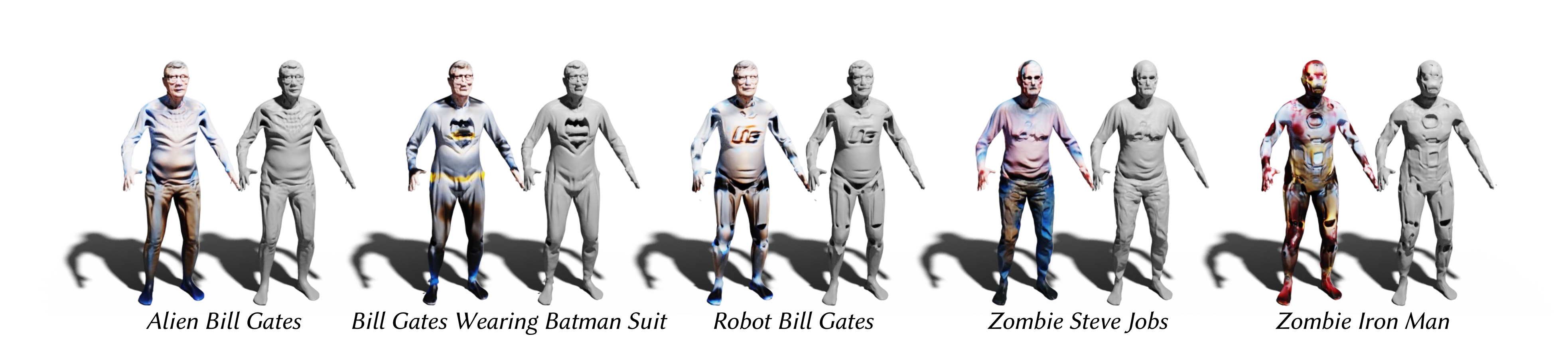}
    \caption{\textbf{Results of Concept Mixing.} We show several concept mixing avatar generation results. Natural blending of different concepts \textcolor{black}{is} demonstrated, which further proves the generalizability of our method. The corresponding input texts are listed below the renderings.}
    \Description{Results of Concept Mixing.}
    \label{fig:conceptmixing}
\end{figure*}

%% file: figures/avatar_user_study.tex
\begin{figure}[t]
    \centering
    \includegraphics[width=\linewidth]{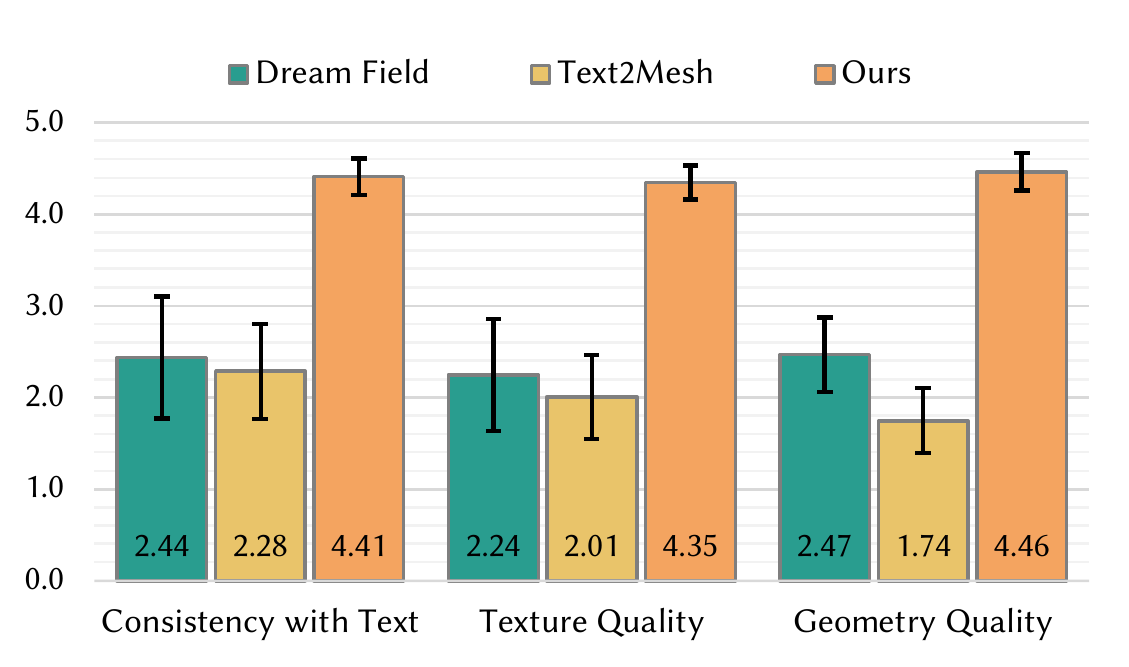}
    \caption{\textbf{User Study on Static Avatar Generation} quantitatively shows the superiority of \nickname{} over other two baseline methods in three aspects: 1) consistency with text, 2) texture quality, and 3) geometry quality.}
    \Description{Results of User Study on Static Avatar Generation.}
    \label{fig:avataruserstudy}
\end{figure}

%% file: figures/motion_user_study.tex
\begin{figure}[t]
    \centering
    \includegraphics[width=\linewidth]{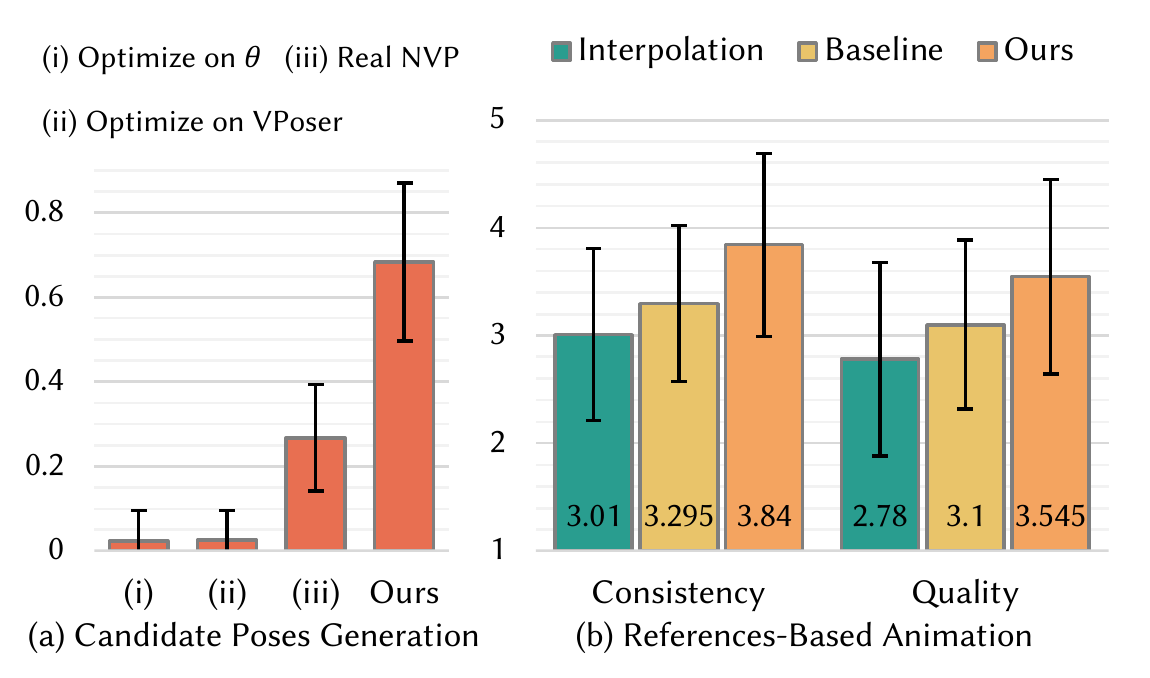}
    \caption{\textbf{Results of User Study on Motion Generation.} For candidate poses generation, our method outperforms both direct optimization and sampling from multi-modality Real NVP. For animation, our method acquires higher scores than both baseline methods noticeably.}
    \Description{Results of User Study on Motion Generation.}
    \label{fig:motionuserstudy}
\end{figure}

%% file: figures/avatar_results.tex
\begin{figure*}[ht]
    \centering
    \includegraphics[width=\linewidth]{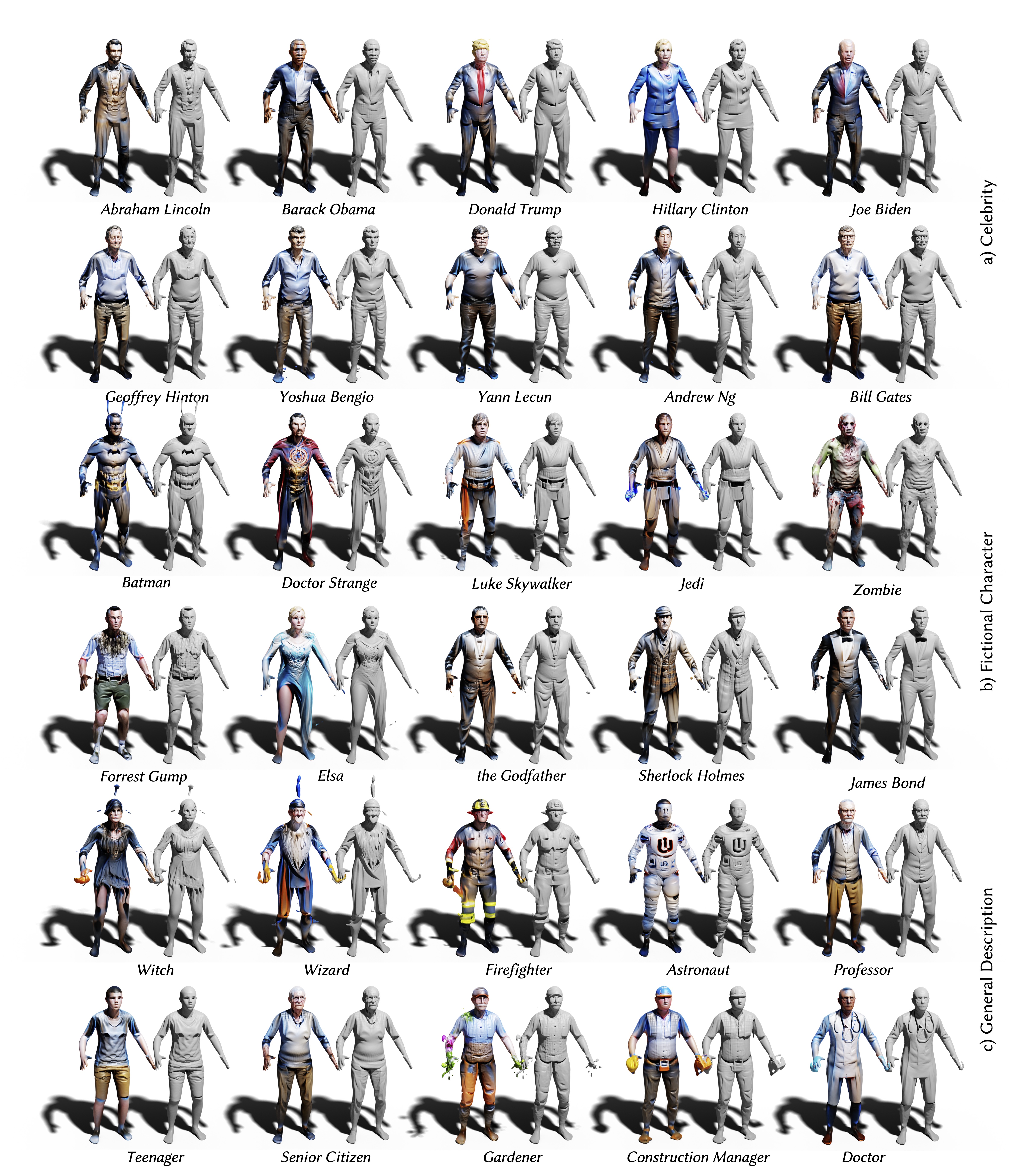}
    \caption{\textbf{Results of Static Avatar Generation.} A broad range of driving texts are tested, including celebrity names, frictional character names and general descriptions. The generated avatars are rendered with and without texture for the convenience of observing textures and geometry.
    }
    \Description{Results of Static Avatar Generation.}
    \label{fig:avatar_results}
\end{figure*}

%% file: figures/morecontrol.tex
\begin{figure*}[ht]
    \centering
    \includegraphics[width=0.9\linewidth]{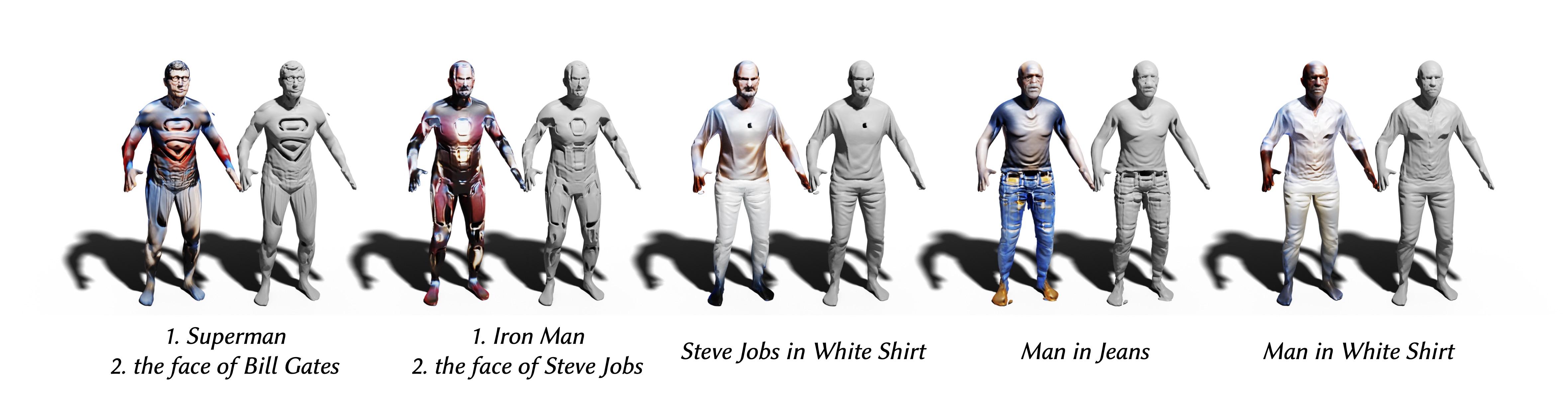}
    \caption{\textbf{Results of More Detailed Controlling.} The left two examples demonstrate the first type of detailed controlling, where the semantic-aware prompt augmentation is adapted. The right three examples further show controls over clothes simply by specifying in the input texts.}
    \Description{Results of More Detailed Controlling.}
    \label{fig:moreconstrol}
\end{figure*}

%% file: figures/geometrycompare.tex
\begin{figure*}[ht]
    \centering
    \includegraphics[width=0.9\linewidth]{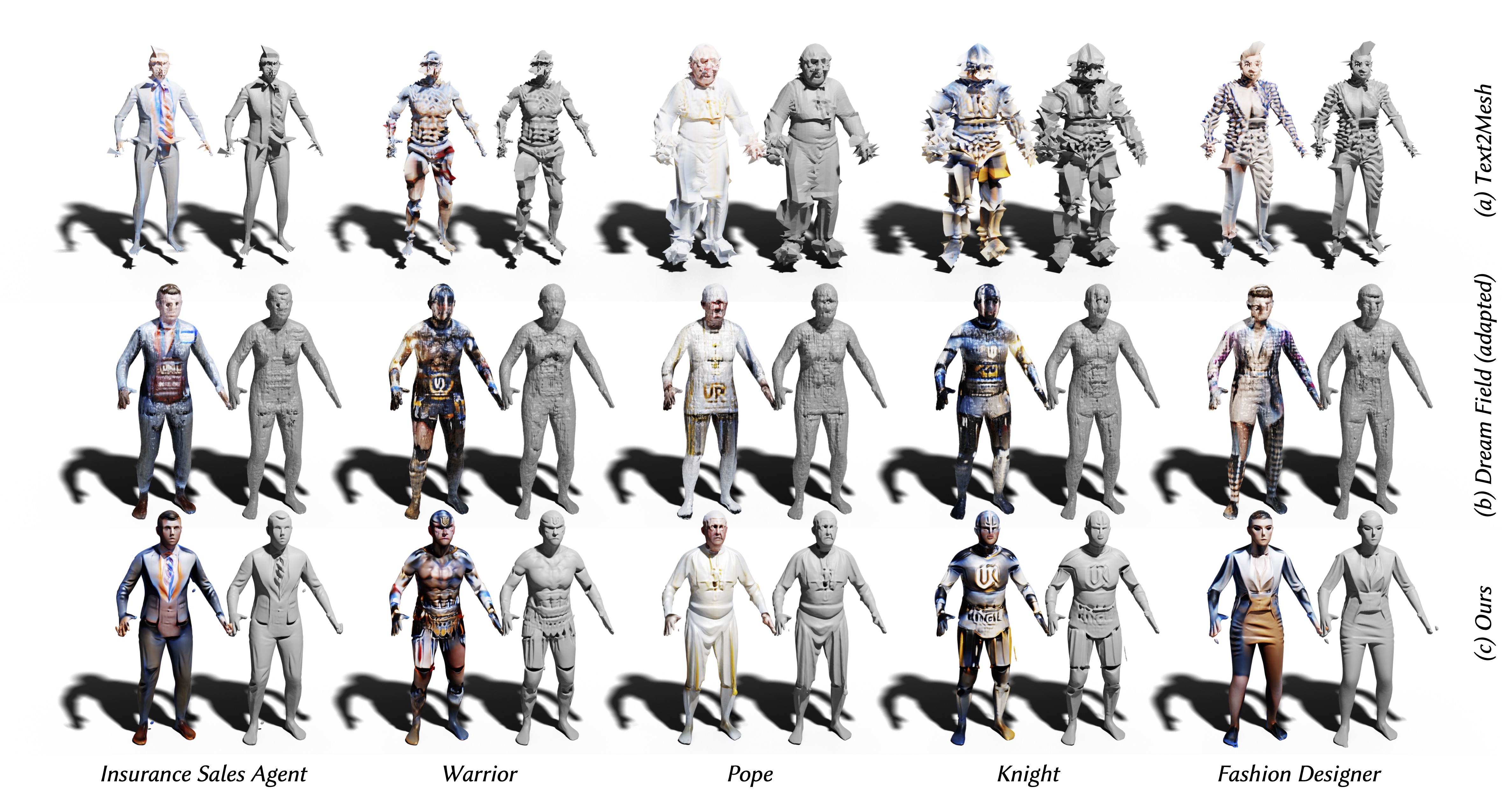}
    \caption{\textbf{Qualitative Comparison with Baseline Methods.} Side-by-side comparisons between our method and (a) Text2Mesh~\cite{michel2021text2mesh} (the first line), (b) Dream Field~\cite{jain2021zero} (the second line) are demonstrated. Results of our method clearly \textcolor{black}{show} better quality in terms of both geometry and texture.}
    \Description{Comparison with Baseline Methods.}
    \label{fig:geometrycompare}
\end{figure*}

%% file: figures/robustness.tex
\begin{figure*}[ht]
    \centering
    \includegraphics[width=0.9\linewidth]{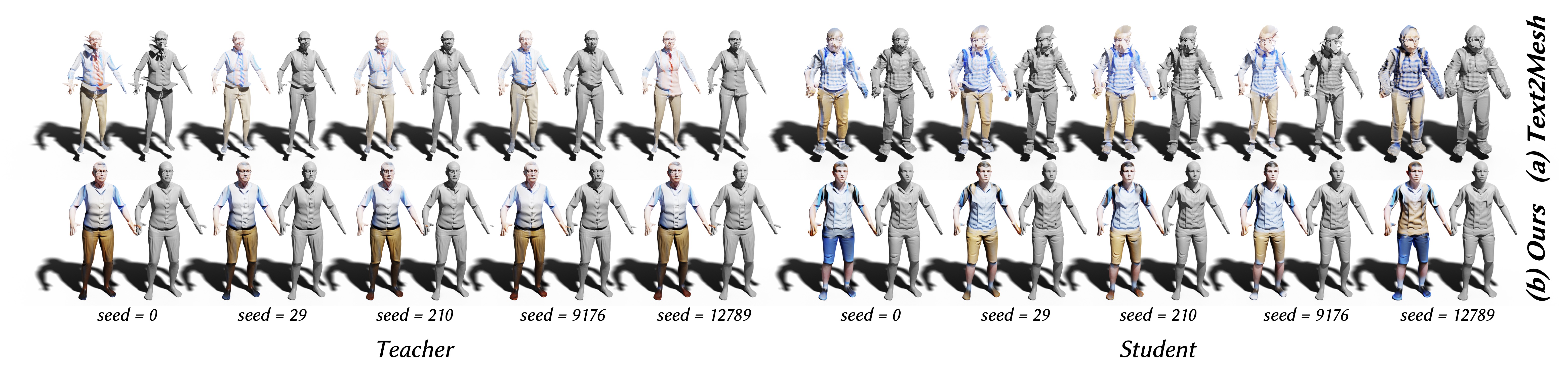}
    \caption{\textbf{Results of Multiple Runs with Different Random Seeds.} By running the optimization multiple times with \textcolor{black}{different} random seeds, we observe that our method succeeds all runs, while Text2Mesh shows unstable results.}
    \Description{Multiple Runs with Different Random Seeds.}
    \label{fig:robustness}
\end{figure*}

%% file: figures/motion_ablation.tex
\begin{figure*}[ht]
    \centering
    \includegraphics[width=0.9\linewidth]{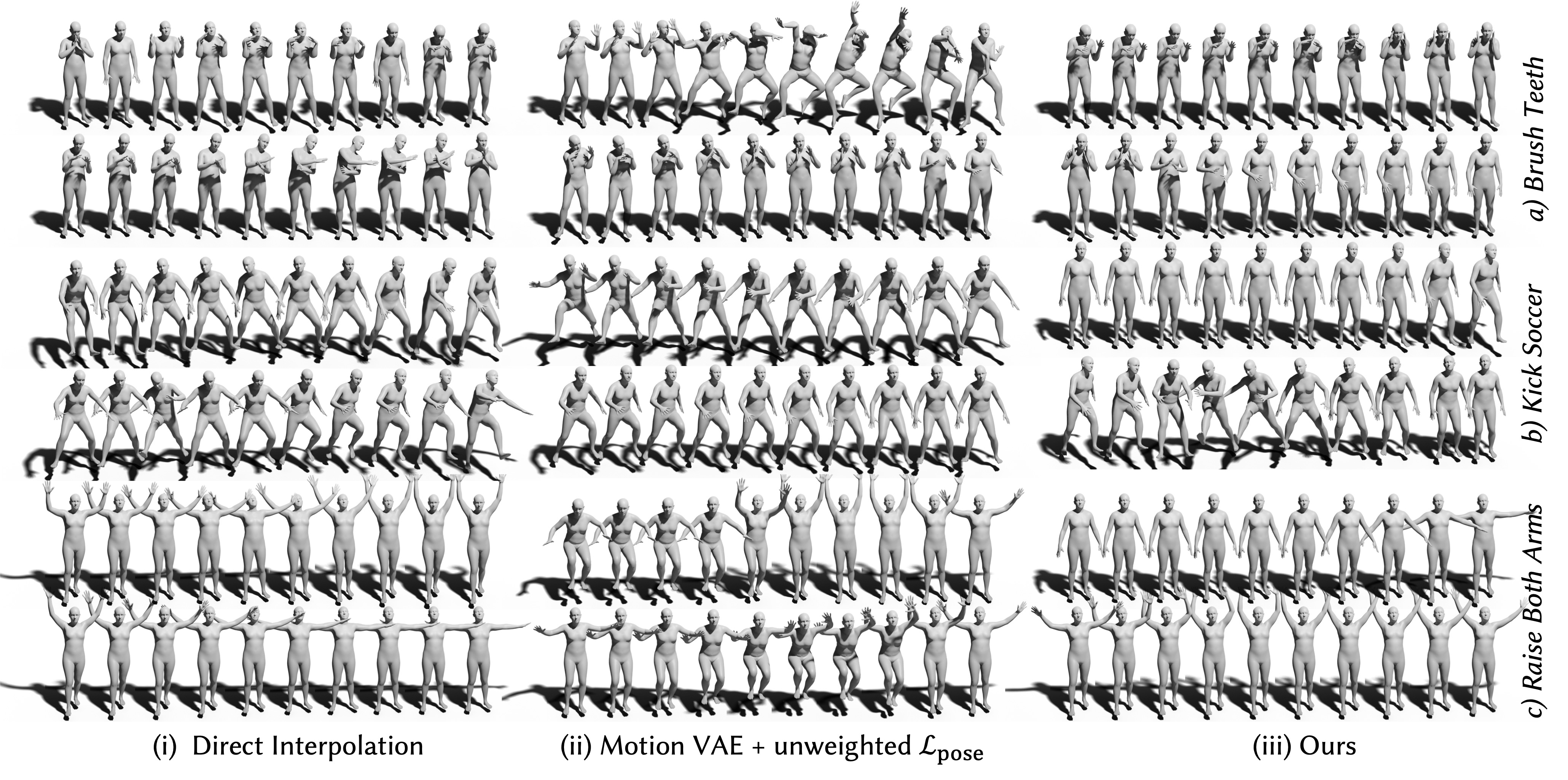}
    \caption{\textbf{Qualitative Ablation of Motion Generation.} Compared with direct interpolation or solely using unweighted $\mathcal{L}_{\textrm{pose}}$, our proposed method can generate stable and reasonable motion sequences that are consistent with the given description.}
    \Description{Qualitative Ablation of Motion Generation.}
    \label{fig:motion_ablation}
\end{figure*}

%% file: figures/pose_results.tex
\begin{figure*}[ht]
    \centering
    \includegraphics[width=0.9\linewidth]{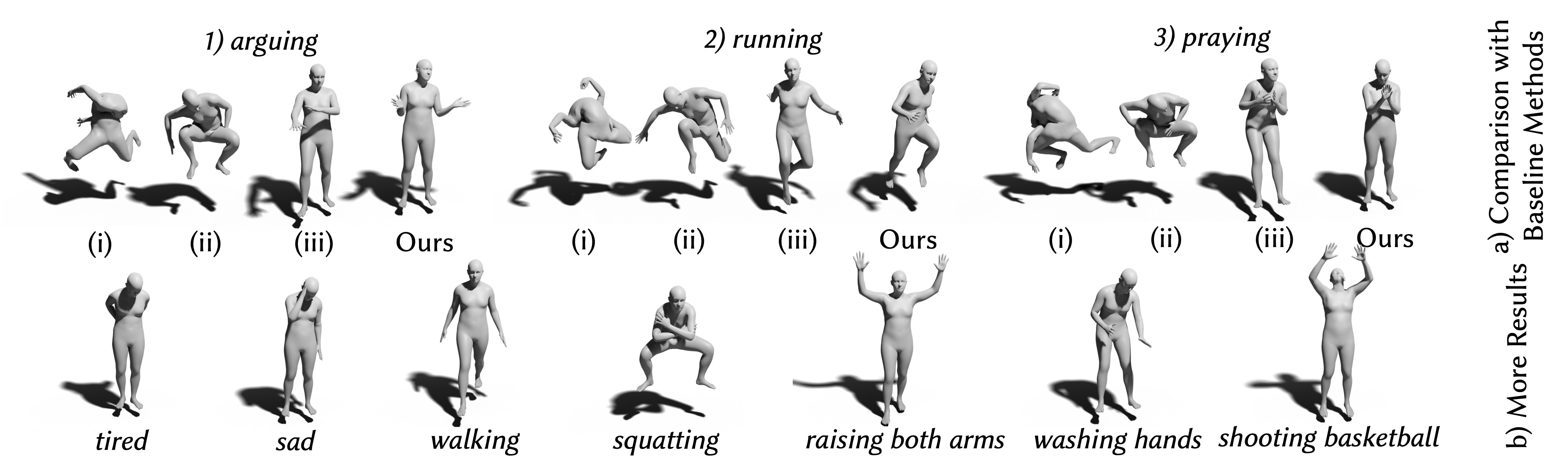}
    \caption{\textbf{Results of the Candidate Pose Generation.} a) compares three baseline methods with ours. Baseline (i) directly optimizes on SMPL parameter $\theta$. Baseline (ii) directly optimizes over the latent space of the VPoser. Both methods can hardly generate reasonable poses. Baseline (iii) utilizes a multi-modality Real NVP,
    which can generate relatively reasonable poses but still worse than our method by a clear margin. b) demonstrates more results of the candidate poses generation.}
    \Description{Results of the Candidate Pose Generation.}
    \label{fig:poseresults}
\end{figure*}

%% file: figures/motion_failure_cases.tex
\begin{figure*}[ht]
    \centering
    \includegraphics[width=0.9\linewidth]{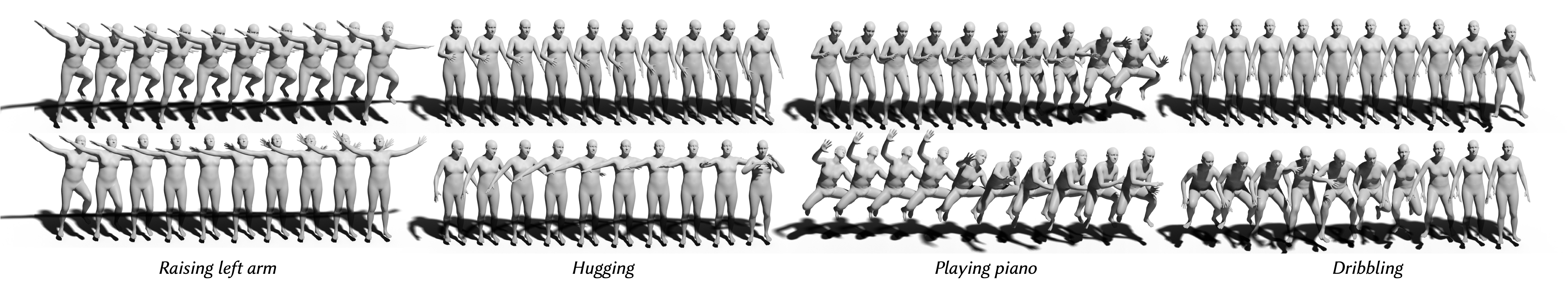}
    \caption{\textbf{Failure Cases of Motion Generation.} More complex motions and more detailed controls over the motions are challenging to generate. Corresponding input texts are listed below.}
    \Description{Failure Cases of Motion Synthesis.}
    \label{fig:motion_failure_case}
\end{figure*}

%% file: sections/05_conclusion.tex
\section{Discussion}
In this work, by proposing \nickname{}, we make the originally complex and demanding 3D avatar creation accessible to layman users in a text-driven style.
It is made possible with the powerful priors provided by the pre-trained models including the shape/ motion VAE and the large-scale vision-language pre-trained model CLIP. Extensive experiments are conducted to validate the effectiveness of careful designs of our methods.

\paragraph{Limitations.} For the avatar generation, limited by the weak supervision and low resolution of CLIP, the results are not perfect if zoomed in. Besides, it is hard to generate avatars with large variations given the same prompt. For the same prompt, the CLIP text feature is always the same. Therefore, the optimization directions are the same, which leads to similar results across different runs. For the motion synthesis, limited by the code-book design, it is hard to generate out-of-distribution candidate poses, which limits the ability to generate complex motions. Moreover, due to the lack of video CLIP, it is difficult to generate stylized motion.

\paragraph{Potential Negative Impacts.}
The usage of pre-trained models might induce ethical issues. For example, if we let $t_\textrm{app}$ = `doctor', the generated avatar is male. If $t_{\textrm{app}}$ = `nurse', the generated avatar is female, which demonstrates the gender bias. We think the problem originates from the large-scale internet data used for CLIP training, which might be biased if not carefully reviewed. Future works regarding the ethical issues of large-scale pre-trained models are required for the zero-shot techniques safe to be used. \textcolor{black}{Moreover, with the democratization of producing avatars and animations, users can easily produce fake videos of celebrities, which might be misused and cause negative social impacts.}